\documentclass[10pt, conference, compsocconf]{IEEEtran}

\ifCLASSINFOpdf

\else

\fi

\usepackage{amsmath,amssymb,amsfonts,epsfig}
\usepackage{algorithm}
\usepackage[noend]{algpseudocode}
\usepackage{mathtools}
\usepackage{hhline}

\usepackage{array}
\usepackage{url}
\usepackage{graphicx}
\usepackage{caption}
\usepackage{subcaption}
\usepackage{multirow}

\usepackage{booktabs} 
\makeatletter
\def\BState{\State\hskip-\ALG@thistlm}
\makeatother

\begin{document}

\title{Optimizing Taxi Carpool Policies via Reinforcement Learning and \\ Spatio-Temporal Mining}

\author{\IEEEauthorblockN{Ishan Jindal\textsuperscript{a}\thanks{\noindent\textsuperscript{a}Work done during an internship at DiDi Research America}\IEEEauthorrefmark{1},
Zhiwei (Tony) Qin\IEEEauthorrefmark{2},
Xuewen Chen\IEEEauthorrefmark{1}, 
Matthew Nokleby\IEEEauthorrefmark{1} and
Jieping Ye\IEEEauthorrefmark{3}}
\IEEEauthorblockA{\IEEEauthorrefmark{1}
Wayne State University, Detroit, MI \qquad \{ishan.jindal, 
xwchen, matthew.nokleby\}@wayne.edu}
\IEEEauthorblockA{\IEEEauthorrefmark{2}DiDi Research America, Mountain View, CA \qquad qinzhiwei@didichuxing.com}
\IEEEauthorblockA{\IEEEauthorrefmark{3}DiDi Research, Beijing, China \qquad yejieping@didichuxing.com}
}

% 

\IEEEoverridecommandlockouts

\maketitle

\begin{abstract}
In this paper, we develop a reinforcement learning (RL) based system to learn an effective policy for carpooling that maximizes transportation efficiency so that fewer cars are required to fulfill the given amount of trip demand. For this purpose, first, we develop a deep neural network model, called ST-NN (Spatio-Temporal Neural Network), to predict taxi trip time from the raw GPS trip data. Secondly, we develop a carpooling simulation environment for RL training, with the output of ST-NN and using the NYC taxi trip dataset. In order to maximize transportation efficiency and minimize traffic congestion, we choose the effective distance covered by the driver on a carpool trip as the reward. Therefore, the more effective distance a driver achieves over a trip (i.e. to satisfy more trip demand) the higher the efficiency and the less will be the traffic congestion. We compared the performance of RL learned policy to a fixed policy (which always accepts carpool) as a baseline and obtained promising results that are interpretable and demonstrate the advantage of our RL approach. We also compare the performance of ST-NN to that of state-of-the-art travel time estimation methods and observe that ST-NN significantly improves the prediction performance and is more robust to outliers.
\end{abstract}


% \IEEEpeerreviewmaketitle

%!Tex root = fullpaper.tex

\section{Introduction}
\label{sec:intro}
In rapidly expanding metropolitan cities, taxis (which include cars working with ride-sharing platforms such as Uber, Lyft and DiDi) play a vital role in residents' daily commute among all the available modes of transportation \cite{schaller2006new}. Based on a survey in NYC \cite{Taxi2011new}, there is a stable demand of taxis, by $666,000$ passengers per day, which is fulfilled by more than $13,000$ taxis in the region. For these expanding cities, to meet the increasing demand of taxis, an emerging problem is to efficiently utilize the existing road networks to reduce potential traffic congestions and to optimize the effective travel time and distance. One promising solution to this problem is \emph{taxi carpool service} \cite{zhang2016carpooling}. In recent years, due to the advancement in the data-driven technologies and availability of big data, it becomes possible to develop more advanced algorithms to solve difficult problems such as taxi travel time estimation \cite{Ishan2017Traveltime,wang2018learning,li2018multi}, taxi routing \cite{han2016routing,verma2017augmenting} etc.

Carpooling is a quick and convenient way to minimize traffic congestion, to reduce air pollution, to save on gas and of course to save on travelers' money. We consider the decision problem in a centralized carpooling service, where carpool assignments are issued to taxi drivers from a central decision system.  In typical carpooling context, a taxi picks up multiple passengers (heading in a similar direction) and proceeds to each of the destinations one-by-one in an efficient manner. Usually, taxis roam around with no passenger on-board and the requests are assigned to a taxi which is in close proximity to the requests. Request assignment to a taxi is a very crucial part of a carpooling service because a bad request assignment might lead a taxi to an area where taxi calls are less frequent and might end up the taxi roaming with no passenger on-board. This kind of situations not only reduce transportation efficiency but cause revenue loss to taxi drivers as well. 

A crucial point to consider in optimizing a carpooling policy is to gauge the future prospect of being able to pick up additional passengers along the way at each decision point.  \textit{Reinforcement Learning (RL)} is a data-driven approach for solving a \textit{Markov decision process (MDP)}, which models a multi-stage sequential decision-making process with a long optimization horizon.
We develop a framework powered by RL to generate data-driven carpooling policy which tells the driver when to accept a carpool request in order to maximize long-term transportation efficiency and reduce traffic congestion. To generate the samples of experience for RL we develop a carpooling simulator which returns a reward and new state corresponding to a state-action pair. 

A key piece of information required to build a carpooling simulation environment is the estimated travel time. Accurate estimates of travel time also help in building intelligent transportation systems such as for developing the efficient navigation systems, for better route planning and for identifying key bottlenecks in traffic networks. The travel time and distance prediction depends heavily on the observable daily and weekly traffic patterns and also on the time-varying features such as weather conditions and traffic incidents. For instance, bad weather or an accident on road slows down the speed of the vehicles and causes lengthy travel time. To tackle this problem, we propose a Spatio-Temporal Neural Network (ST-NN) approach which jointly learns the travel time and the travel distance from the raw GPS coordinates of an origin, a destination and the time-of-the-day. The beauty of the ST-NN is that it does not require any sort of feature engineering. We then use these estimates to optimize carpooling in simulation to minimize the extra travel time traveled by each of the passengers on board. Throughout the paper, we denote vectors as bold case letters.

\textbf{Summary of results}: We summarize our main technical contributions as follows:
\begin{enumerate}
\item In Section \ref{sec:testbed}, we develop a carpooling simulation environment using NYC taxi trip dataset to generate training experiences for the RL framework.
\item In Section \ref{sec:st-nn}, we develop a deep neural network model (ST-NN) which predicts the  travel time and distance directly from the GPS coordinates of origin and destination locations in the city, without building any route or map between the locations.
\item In Section \ref{sec:q-learning}, we present an RL framework to obtain a data-driven carpooling policy for the taxi drivers in order to maximize the long-term transportation efficiency and reduce traffic congestion based on the time of the day and the day of the week.
\end{enumerate}

The rest of the paper is organized as follows: Section \ref{sec:background} describes the background and related work on carpooling, travel time estimation and provides a brief introduction to reinforcement learning. In Section \ref{sec:problem}, we first describe the problem and the MDP formulation of the problem for reinforcement learning. We empirically evaluate the learned RL policy and the performance of ST-NN approach in Section \ref{sec:performance}, and Section \ref{sec:conclusion} concludes the work with future directions.

% !Tex root = fullpaper.tex

\section{Background and Related Work}
\label{sec:background}

\subsection{Carpooling} 
Carpooling in taxis and ride-share services has been widely studied. At first \cite{fagin1983fair} presented fair share carpool scheduling algorithm. In classical carpool settings, various assumptions were made such as fixed and regular travel path of passengers \cite{fagin1983fair}. \cite{santi2014quantifying} showed that with a small increase in travel time it is possible for almost $80\%$ of the taxi trips in Manhattan to be shared by two riders. \cite{alonso2017demand} considered solutions to ride-sharing which is scalable to a large number of passengers. Other works which explore carpooling involve real-time carpooling on mobile-cloud architecture \cite{ma2015real}, social effects of carpooling \cite{meurer2014social}, etc. Our focus is on a data-driven approach to optimize central carpool decision policies through reinforcement learning and building a training simulation environment based on historical taxi trip data.

\subsection{Travel Time Estimation}
Most of the studies in literature for travel time estimation are focused on predicting the travel time for a sequence of locations for a fixed route. Commonly used techniques include (1) estimating travel time using historical trips; (2) using real time road speed information \cite{narayanan2015travel}. The two common approaches for route travel time estimation include \emph{segment-based approaches} and \emph{path-based approaches}. In a segment-based approach, travel time is estimated on links (straight subsections of a travel path with no intersections) first and then summed up to estimate the overall travel time. The link travel time is generally calculated by using loop detector data and floating car data \cite{kesting2013traffic} \cite{zhan2013urban}. Whereas in floating car data, GPS enabled cars are used to collect timestamped GPS coordinates. The available dataset, in ST-NN, can think of as the special case of floating car data, where only the origin and destination GPS coordinates are recorded.

One of the major drawbacks of the segment-based approach is that it is unable to capture the waiting times of a vehicle at the traffic lights, which is a very important factor for estimating the accurate travel time. Therefore, some methods are developed to consider the waiting time at the intersections for travel time estimation \cite{li2015inferring}. In path-based methods, sub-paths (links + waiting time at intersections) are concatenated to predict the most accurate travel time \cite{hofleitner2012learning}. Our method is the special case of the path-based method, where sub-path is the entire path from origin to destination containing information about the waiting times at all the intersections. In addition to these methods, \cite{morgul2013commercial} proposes a neighbor-based method for travel time estimation by averaging the travel time for all the samples in training data having the same origin, destination, and time-of-day.

In this paper, we jointly predict the travel time and distance from an origin to a destination as a function of the time-of-day using the historical NYC travel trips data. Since the available NYC taxi trip dataset does not contain full trip trajectories, we treat it as a full path travel time estimation problem. As an alternative solution for travel time estimation, one can first find the specific trajectory path (route) between origin and destination and then estimate the travel time for that route \cite{yuan2010t}. Although, obtaining the travel route information is important, we can think of a certain real scenario where route information is not as much important as travel time. For example, the travel route is of much less concern than the travel time to a non-driving taxi passenger.

\subsection{Reinforcement Learning} 
Reinforcement learning (RL) is used for learning an optimal policy in a dynamic environment. In RL, an agent observes a state $\mathbf{S}_0$, takes an action $a$ in the state $\mathbf{S}_0$, receives a reward $r$ from the environment, transitions to the next state $\mathbf{S}_1$ and keeps repeating this procedure until it reaches a terminal state which ends the episode. Initially, the agent randomly picks an action from the action space given a state because the agent has no knowledge of which action has to be taken in a given state. This means that the agent is \emph{exploring} its environment by taking random actions. As the time proceeds, the agent gets more confidence on its predicted actions and starts \emph{exploiting} its knowledge by taking an action with highest estimated value and produce the greatest reward. In RL, trade-off between exploration and exploitation is crucial. 

RL methods can be broadly divided into model-based and model-free learning methods. The model consists of the knowledge of the environment: the state transition probabilities and the reward function. In both types of methods, the model is not known in advance. In model-based RL methods, the transition model is first learned and then used to derive an optimal policy \cite{chakraborty2011structure}. Learning a model requires exhaustive exploration which is very costly for a large state space. However, it is possible to learn an optimal policy without even knowing the model using model-free RL methods such as temporal difference learning and Monte Carlo methods \cite{sutton1998reinforcement}. RL is primarily concerned with these model-free methods where an optimal policy is learned from the samples of experience obtained from interacting with the environment. Model-free methods often require a large number of experiences to learn an optimal policy. In this work, we develop a carpooling simulator which generates a lot of experiences for model-free RL methods. 

Q-Learning \cite{watkins1992q} is a widely used model-free RL method because of its computational simplicity. The simplest method to obtain a policy is tabular Q-learning where the algorithm keeps a record of the value function in a tabular form \cite{sutton1998reinforcement}. However, when the state and/or action space is large, maintaining such a big table is expensive and is sometimes even infeasible. Therefore, function approximation techniques are used to approximately learn this table. For example, deep RL methods use deep neural networks to approximate the Q-value function (Deep Q-Networks (DQN)) \cite{mnih2015human}. Deep RL has become popular because of its success in playing games \cite{silver2016mastering,mnih2013playing} where the state space has hundreds of features. In carpooling, the state space is huge, as the state is composed of latitude and longitude coordinates along with a continuous variable --- time of day. Therefore, DQN is suitable in this problem for generating an optimal policy.

RL methods have been used for routing autonomous taxis \cite{han2016routing}, where taxis are assigned routes with the highest probability of finding a passenger and also in advising the taxis' about the right location where they can find more customers and hence maximize their income \cite{verma2017augmenting}. 
% !Tex root = fullpaper.tex

\section{Problem Definition}
\label{sec:problem}

In this section, we set the basic terminology, Markov Decision Process (MDP) formulation and the travel time estimation problem that is integral to building the simulation environment for reinforcement learning.

\subsection{Data Mapping}
\label{subsetion:mapping}
\begin{figure}[h]
    \centering
    \begin{subfigure}[t]{0.5\columnwidth}
        \centering
        \includegraphics[width = \columnwidth]{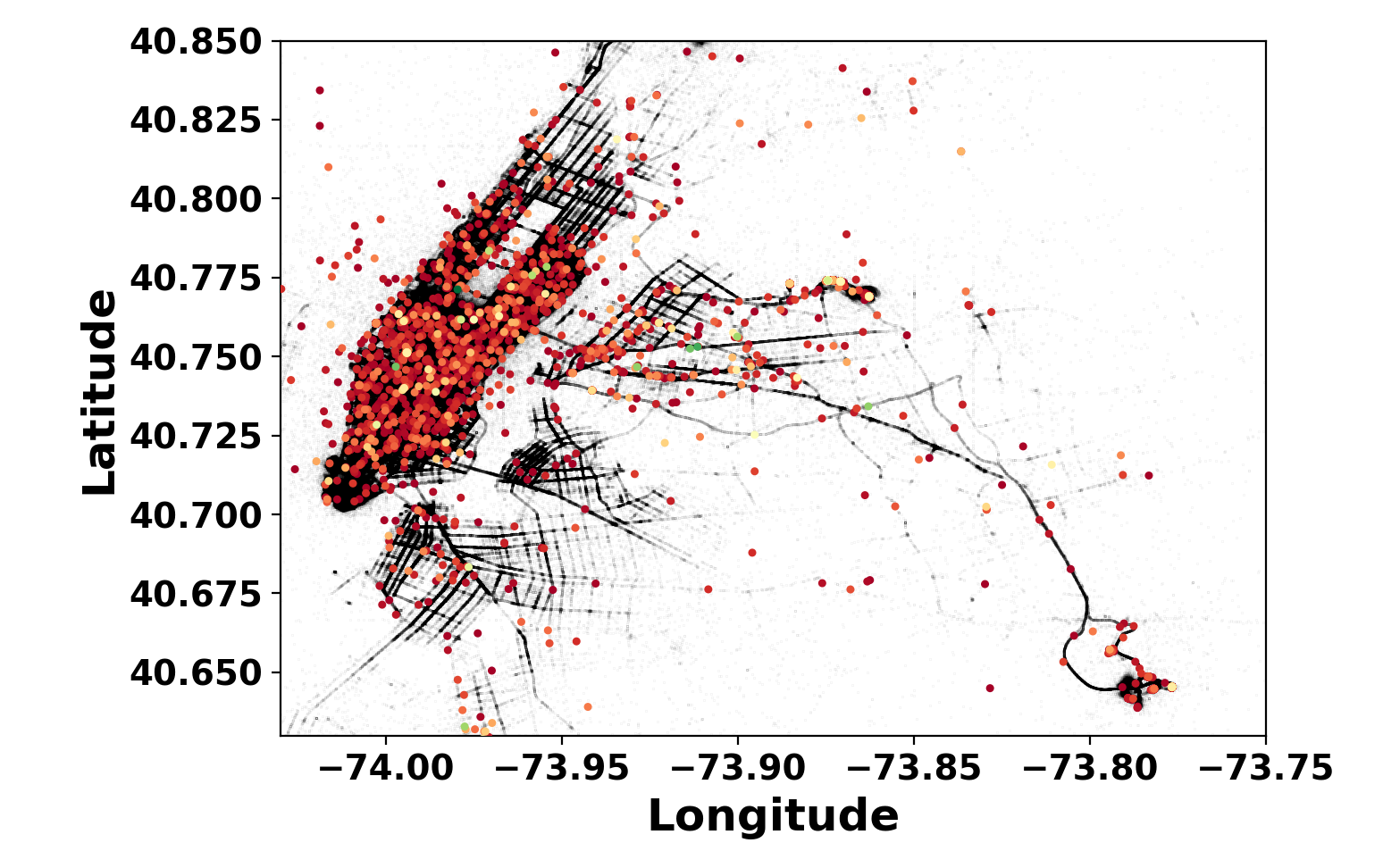}
        \caption{}
        \label{fig::PickDist}
    \end{subfigure}%
    ~ 
    \begin{subfigure}[t]{0.5\columnwidth}
        \centering
        \includegraphics[width = \columnwidth]{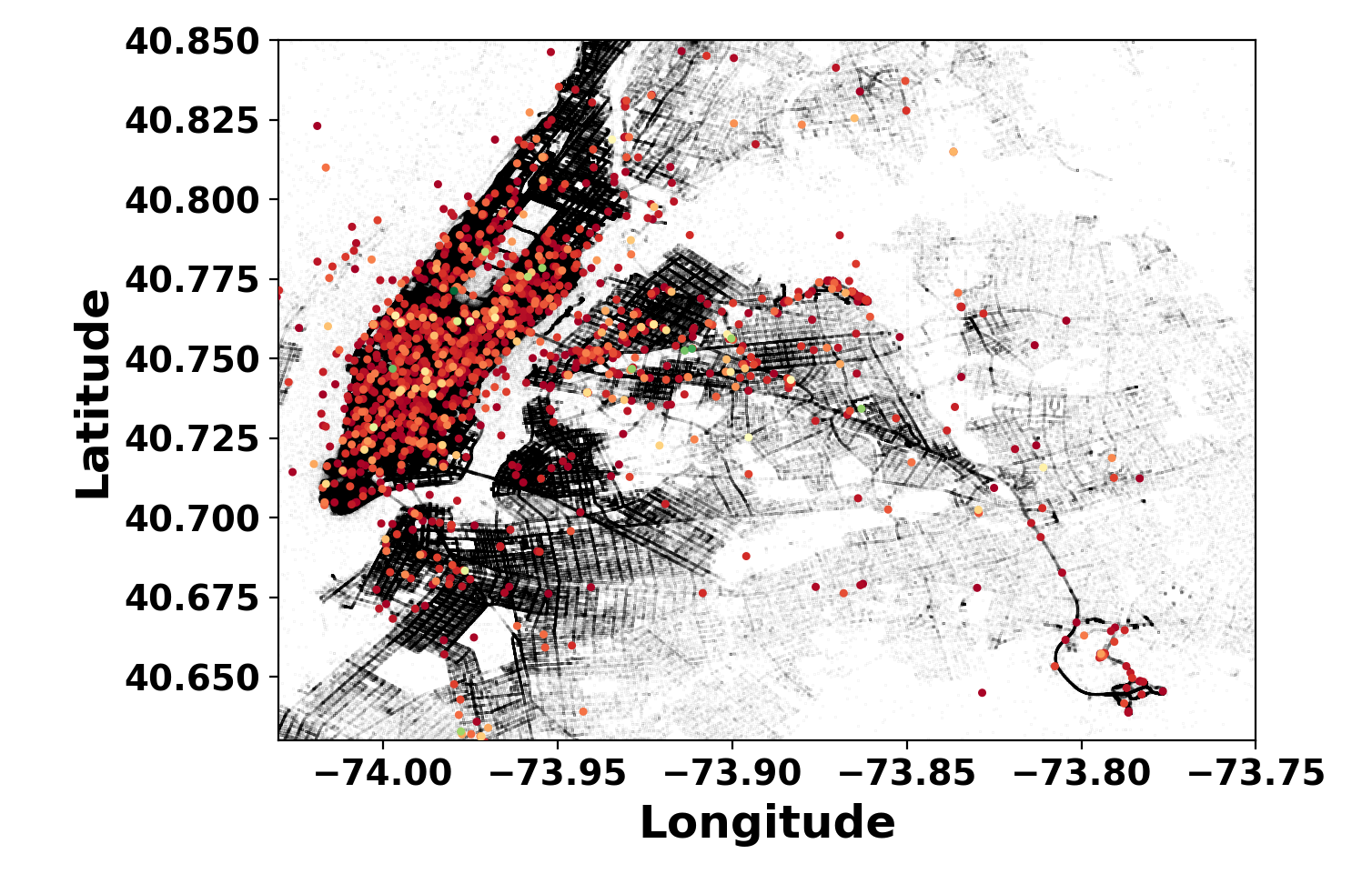}
        \caption{}
        \label{fig::DropDist}
    \end{subfigure}
    \caption{NYC GPS Coordinates Distribution (a) Taxi pickup; (b) Taxi dropoff}
    \label{GPS}
\end{figure}

A publicly available gigantic taxi trip dataset contains $173$M taxi trips for the New York City during the year 2013 \cite{dataURL}. This dataset describes every single trip by 21 different variables. Fig. \ref{GPS} outline the provided GPS coordinates where Fig. \ref{fig::PickDist} and \ref{fig::DropDist} show the density of pickup and drop-off GPS coordinates, respectively. 

Geo-coordinates are continuous variables. In the urban cities like NYC, because of tall buildings and dense areas, it is quite possible to get the erroneous GPS coordinates while reporting the data. Other sources of erroneous recording of GPS coordinates include atmospheric effects, multi-path effects and clock errors. For more information, we refer the reader to \cite{grewal2007global}. Therefore, to combat the uncertainties in GPS recording, a data preprocessing step is needed to process the raw GPS data. Hence, we discretized the GPS coordinate into 2-D square cells, say of $200m$ longitude and $200m$ latitude. All the GPS coordinates of a square cell are represented by the lower left corner of that square cell.

Similar to location mapping, we also discretized the time-of-day as a 1-D time cell. From the NYC dataset, we observe that the average travel time of a taxi for weekday differs from the weekend. Therefore, we differentiate the time-of-day of weekdays from weekends. The time-of-day of the weekend is incremented by $3600*24$ seconds of time-of-day of the weekday. For a time cell of $10$ minutes we obtain a total $288$ time cells.

%!TEX root = fullpaper.tex

\subsection{MDP Formulation}\label{sec:mdp_formulation}
We model the carpooling problem from a driver's perspective through the following MDP.

\begin{figure}[t]
\centering
\includegraphics[width=\columnwidth]{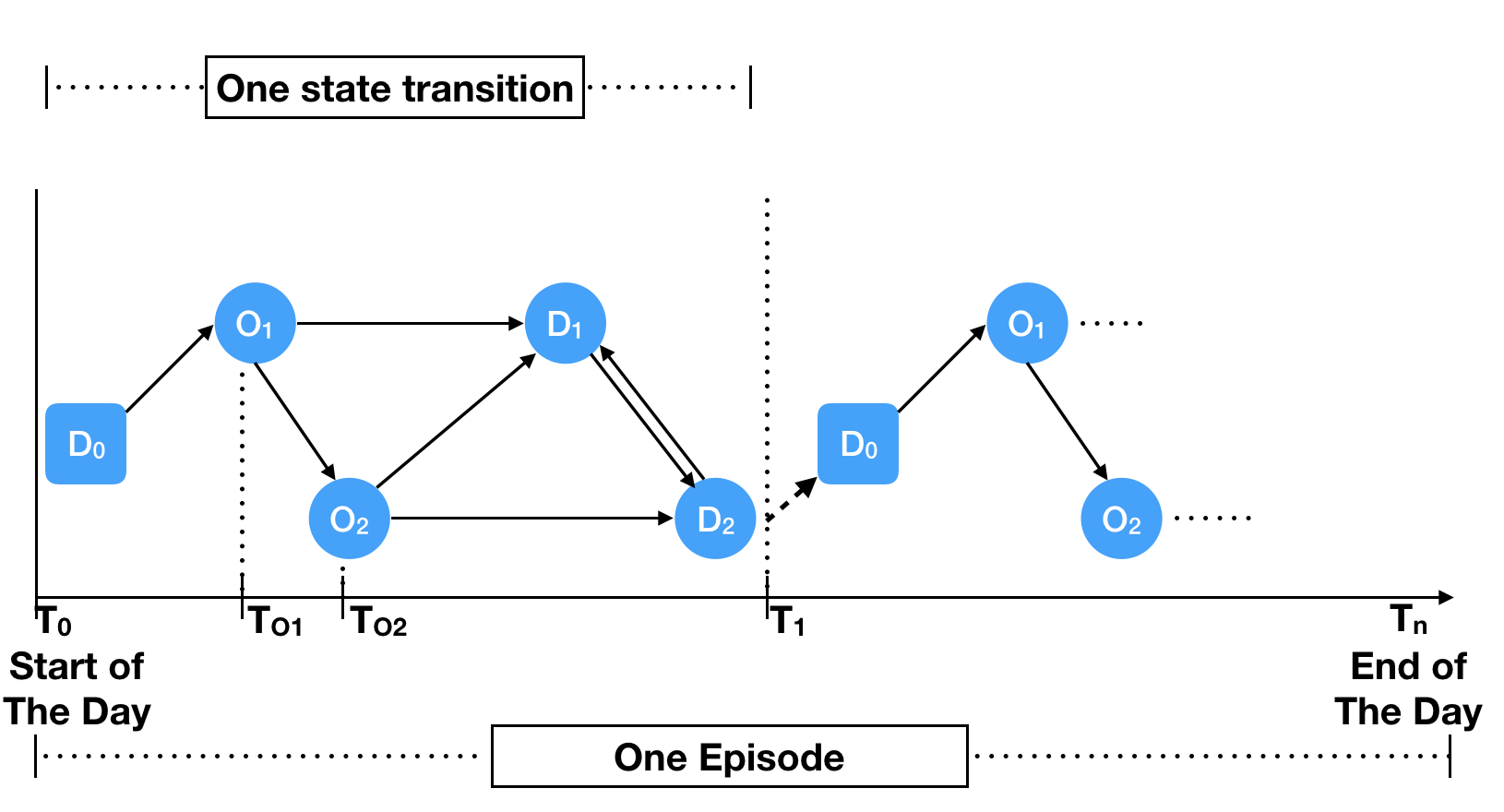}
\caption{State Transition Process}
\label{fig:simulator}
  \vspace{-0.1in}
\end{figure}

% \begin{itemize}
\textbf{State}, $\mathbf{S}_i := (l_i,t_i)$: represents the $i$-th state of an agent (taxi). Here, $l_i$ is 2-D tuple, represents the GPS coordinates $(\mathrm{Lat}_i, \mathrm{Lon}_i)$, and $t_i$ denotes the time of the day in seconds. One should not confuse the state of the taxi with the actual origin of a taxi trip. State of the taxi can be different from the origin of a trip. \\
	% We differentiated the time of the day for weekday and weekend. \\
\textbf{Action,} $a := (W, TK1, TK2)$: Here, $W$ represents the \emph{wait} action, $TK1$ is the action of assigning single passenger that is non-carpool, and we call it \emph{take one} action and $TK2$ corresponds to \emph{carpool}. Any one of the actions from this set of actions can be assigned to a taxi. In this work, we assume that at most two taxi calls can be assigned to a taxi. As we will explain later, $a$ corresponds to the top-level carpool decisions. We leave the low-level trips assignment to the environment in this work.\\
\textbf{Reward,} $r$: We define the reward as the effective distance traveled by the taxi throughout a transition. The effective distance of any trip is defined as the sum of actual distances between the origin of the trips to the destination of the individual trips, obtained from the historical dataset. For instance, when $TK1$ action is assigned to taxi, the effective distance is the actual distance between origin and destination of the trip whereas in $W$ action, the effective distance is always zero.  A good $TK2$ action will yield an effective trip distance longer than the distance actually traveled by taxi.

We choose effective distance as a reward because for a fixed interval of time if all the taxis can cover more effective distance then a large demand of taxi rides can be fulfilled by a few number of drivers on the road. In general, the ideal situation is that through carpooling, the entire group of drivers can cover more effective trip distance than they actually travel, thus reducing the possibility of traffic congestion.\\
\textbf{Episode:} One episode is one complete day, from 0:00 AM to 23:59 PM. Hence, one episode completes when the $t$ component of the state of the taxi reaches 23:59 PM.\\
\textbf{State Transition:} When a taxi completes an assigned action, the state of the taxi gets updated and this change in the state is termed as \emph{state transition}. In Fig. \ref{fig:simulator}, $T_0 \rightarrow T_1 $ is one state transition and $T_0 \rightarrow T_n $ defines one episode. Here, $T_0$ denotes the start of the day, $T_1$ denotes the completion of one state transition and $T_n$ represents the end of the day or end of the episode. These state transitions continue until reaching the termination state. After a state transition the taxi can either pause for a while (driver wants to take rest) or opens to other assignments immediately, but we assume that the taxi is willing to get the assignments all the time.

\subsection{Travel Time Estimation}
We define the travel time as the time taken by a vehicle in moving from one location to another. Similarly, travel distance is defined as the distance transversed by a vehicle between two locations. In simple words, one can think of ST-NN as to estimate the travel distance and time between an origin $(o)$ and a destination $(d)$ at a particular time $(t)$ time-of-day.

We define a taxi trip $\mathbf{p}_i$, as a 5-tuple $(\mathbf{O}_i, \mathbf{D}_i, t_i, d_i, T_i)$, starting from the origin $\mathbf{O}_i$ at time-of-day $t_i$ heading to the destination $\mathbf{D}_i$, where $d_i$, is the travel distance and $T_i$ represents the travel time. Both the origin and destination are 2-tuple GPS coordinates, that is $\mathbf{O}_i = (\mathrm{Lat}_i, \mathrm{Lon}_i)$ and $\mathbf{D}_i = (\mathrm{Lat}_i, \mathrm{Lon}_i)$, and time-of-day ($t_i$) is in seconds. An intuitive reason to include time-of-day $t_i$ as part of a taxi trip is due to different traffic conditions at different times. For example, one can encounter heavy traffic at peak hours than off-peak hours. The traffic patterns on weekdays is also different from weekends. Similar to \cite{wang2016simple}, we assume that the intermediate location or travel trajectory is not known and only the end locations are available. We define a query $\mathbf{q}_i$ as a pair $($\emph{origin, destination, time-of-day}$)_i $  input to the system and corresponding pair $($\emph{travel time, travel distance}$)_i$ as an output. Therefore, for the travel time estimation, the only input query is  $(\mathbf{O}_i, \mathbf{D}_i, t_i)$, and the network estimates $(d_i, T_i)$. Given the historical database of $N$ taxi trips $\mathcal{X} = \{\mathbf{p}_i\}_{i=1}^N$, our goal is to estimate the travel distance and time, $(d_q, T_q)$ for a query $\mathbf{q} = (\mathbf{O}_q, \mathbf{D}_q, t_q)$.

\section{Training environment: Carpooling Simulator}
\label{sec:testbed}

In order to train an RL agent that makes optimal carpooling decisions, we develop a carpooling simulation environment from a single taxi driver's perspective, corresponding to our MDP formulation. In our training environment, the transition dynamics is divided into two levels: the action space defined  in Section \ref{sec:mdp_formulation} and the more granular decision on trips assignment.
We assume that the system performs the decision-making at both levels for the taxi driver. The RL learned policy makes only the first-level decisions (assigning an action to the taxi which maximizes the long term transportation efficiency) whereas the secondary decisions are determined by a fixed algorithm described below in this section.

From Fig. \ref{fig:simulator}, at the start of the episode, $D_0$ is the initial state $\mathbf{S}_0 =  (l_0,t_0)$ of the taxi, this should not be confused with the actual origin of the taxi trip which is $O_1$. $\mathbf{s}_{O_1} = (l_{O_1}, t_{O_1})$ is the intermediate state of the taxi when it picks up the first passenger. Now, we define all the actions.

\textbf{Wait Action $(W)$}: When a wait action is assigned to the taxi at state $\mathbf{s}_0 = (l_0,t_0)$, taxi stays at the current location $l_0$ while the time $t_0$ advances by $t_d$, where $t_d$ is the delay time. Therefore, the next state of the driver would be $(l_0, t_0+t_d)$ as described in Algorithm \ref{W_Action}.

\begin{algorithm}
\caption{Wait Action}\label{W_Action}
\begin{algorithmic}[1]
\Procedure{}{}
\BState \textit{Given}: $\mathcal{D}$, $l_0$, $t_0$, $t_d$, $T = 600$ Sec., $T_c$
\State $\textit{$\mathbf{S}_0$} \gets  \textit{$(l_0,t_0)$ } $
\State $\textit{$\mathbf{S}_1$} \gets  \textit{$(l_0, t_0+t_d)$}$, $\textit{$r$} \gets  \textit{$0$ } $
\EndProcedure
\end{algorithmic}
\end{algorithm}

\textbf{Take 1 Action $(TK1)$}: Given the initial state of the taxi $\mathbf{S}_0$ and the $TK1$ action,  
% a taxi trip is assigned to the taxi for which the taxi can reach to the origin of a taxi trip, say $O_1$, in the time less than the pick up time of the passenger from that taxi trip origin. 
% For example, 
at first the taxi trip search area is reduced by finding all the taxi trips having pickup time in the range $t_0$ to $(t_0+T)$ irrespective of the origin of the taxi trips, where $T$ defines the search time window and is fixed, say ten minutes. The taxi trip search area is further reduced by finding all the taxi trips where the taxi can reach before the pickup time from its initial state $\mathbf{S}_0$. If there is no such trip origin, the taxi continues waiting at its current location $l_0$ but the time advances to $t_0+T$ and the state of the taxi becomes $\mathbf{S}_1 = (l_0, t_0+T)$. Whereas, if there exist such taxi trips, then a taxi trip with minimum pick up time is assigned to the taxi. Finally, the taxi picks up the passenger from the origin of taxi trip and drops the passenger at the destination and updates its state to $\mathbf{S}_1 = (l_{D_1}, t_{D_1})$ and completes the state transition. Here, $l_{D_1}, t_{D_1}$ represent the dropoff location and time of first passenger, respectively. Take 1 action is described in Algorithm \ref{TK1_Action}.
\begin{algorithm}
\caption{Take 1 Action}\label{TK1_Action}
\begin{algorithmic}[1]
\Procedure{}{}
\BState \textit{Given}: $\mathcal{D}$, $l_0$, $t_0$, $t_d$, $T = 600$ Sec., $T_c$
\State $\textit{$\mathbf{S}_0$} \gets  \textit{$(l_0,t_0)$ } $
\State $\textit{$\mathcal{D}^\prime$} \gets \textit{$\mathcal{D}(t_0 \leq PickupTime(\mathcal{D}) \leq (t_0+T))$} $
\State $\textit{$\mathcal{D}^\prime$} \gets \textit{$\mathcal{D}^\prime(t(D_0, O_1) \leq PickupTime(\mathcal{D}^\prime) )$} $
\If {$\mathcal{D}^\prime == \{\phi\}$} 
	\State $\textit{$\mathbf{S}_1$} \gets  \textit{$(l_0, t_0+t_d)$}$, $\textit{$r$} \gets  \textit{$0$ } $
\Else 
	\State $\textit{$O_1$} \gets  \textit{$\arg \min_{PickupTime} \mathcal{D}^\prime $ } $
	\State $\textit{$\mathbf{S}_1$} \gets  \textit{$(l_{D_1},t_{D_1})$}$, $\textit{$r$} \gets  \textit{$d(O_1, D_1)$ } $
\EndIf
\EndProcedure
\end{algorithmic}
\end{algorithm}

\textbf{Take 2 Action $(TK2)$}: Now, if the \emph{take 2} action (corresponds to carpool) is assigned to a taxi, given the initial state $\mathbf{S}_0$, first taxi call is assigned to the taxi similar to the $TK1$ action.
% \emph{take 1} action. Once the first taxi call is assigned, the taxi reaches the origin location $O_1$ of the first passenger and updates its intermediate state $\mathbf{s}_{O_1} = (l_{O_1}, t_{O_1})$. From the intermediate state $\mathbf{s}_{O_1}$, how a second taxi call is assigned to the taxi is described in Algorithm \ref{TK2_Action}. 
At this intermediate state $\mathbf{s}_{O_1} = (l_{O_1}, t_{O_1})$, a second taxi call $\mathbf{s}_{O_2} = (l_{O_2}, t_{O_2})$ is assigned to the driver by following the same procedure of assigning the first taxi call. The only difference is the taxi trip's pickup time range. For the second taxi call, the taxi trip search area is reduced by selecting all the taxi calls in pickup time range $t_{O_1}$ to $(t_{O_1}+ (T_c * t(O_1,D_1)))$ irrespective of the origin locations of the taxi trips. This means that the taxi has to wait at the intermediate state $\mathbf{s}_{O_1}$ for $(T_c * t(O_1,D_1))$ seconds while the search for another taxi call is being made. Here, $T_c \in (0,1)$ is an important parameter which controls the taxi trip search area for the second taxi call assignment. 
\begin{algorithm}
\caption{Take 2 Action}\label{TK2_Action}
\begin{algorithmic}[1]
\Procedure{}{}
\BState \textit{Given}: $\mathcal{D}$, $l_0$, $t_0$, $t_d$, $T = 600$ Sec., $T_c$
\State $\textit{$\mathbf{S}_0$} \gets  \textit{$(l_0,t_0)$ } $
\State $\textit{$\mathcal{D}^\prime$} \gets \textit{$\mathcal{D}(t_0 \leq PickupTime(\mathcal{D}) \leq (t_0+T))$} $
\State $\textit{$\mathcal{D}^\prime$} \gets \textit{$\mathcal{D}^\prime(t(D_0, O_1) \leq PickupTime(\mathcal{D}^\prime) )$} $
\If {$\mathcal{D}^\prime == \{\phi\}$} 
	\State $\textit{$\mathbf{S}_1$} \gets  \textit{$(l_0, t_0+t_d)$ }$, $\textit{$r$} \gets  \textit{$0$ } $
\Else 
	\State $\textit{$O_1$} \gets  \textit{$\arg \min_{PickupTime} \mathcal{D}^\prime $ } $
	\State $\textit{$\mathbf{s}_{O_1}$} \gets  \textit{$(l_{O_1},t_{O_1})$ } $ 
	\State $\textit{$T_1$} \gets  \textit{$T_c*(t(O_1, D_1) $ } $

	\State $\textit{$\mathcal{D}$} \gets \textit{$\mathcal{D}(t_{O_1} \leq PickupTime(\mathcal{D}) \leq (t_{O_1}+T_1))$} $
	\State $\textit{$\mathcal{D}$} \gets \textit{$\mathcal{D}(t(O_1, D_1) \leq PickupTime(\mathcal{D}) )$} $
	\If {$\mathcal{D} == \{\phi\}$} 
		\State $\textit{$\mathbf{S}_1$} \gets  \textit{$(l_0, t_0+t_d)$ }$, $\textit{$r$} \gets  \textit{$0$ } $
	{}\Else % \textit{Take 2 $(TK2)$ action}
		\State $\textit{$\mathrm{T_{Ext_I}}$}, \textit{$\mathrm{T_{Ext_{II}}}$} \gets \emph{ETT($\mathcal{D}$, $O_1,D_1$)} $
		\State $\textit{$O_2$} \gets  \textit{$\arg \min  (\mathrm{T_{Ext_I}}+\mathrm{T_{Ext_{II}}})$ } $
		\State $\textit{$\mathbf{s}_{O_2}$} \gets  \textit{$(l_{O_2},t_{O_2})$ } $
		\If {$\mathrm{T_{Ext_I}} < \mathrm{T_{Ext_{II}}}$} 
			\State $\textit{$\mathbf{S}_1$} \gets  \textit{$(l_{D_1}, t_{D_1})$ } $
		\Else
			\State $\textit{$\mathbf{S}_1$} \gets  \textit{$(l_{D_2}, t_{D_2})$ } $	
		\EndIf
		\State $\textit{$r$} \gets  \textit{$d(O_1, D_1) + d(O_2, D_2)$ } $	
	\EndIf
\EndIf
\EndProcedure
\end{algorithmic}
\end{algorithm}

In carpooling scenarios, for first passenger/customer satisfaction, we can't fix the taxi call search area for second taxi call assignments. For instance, let us fix the size of search time window $T = 600s$. Similar to first taxi call. The pickup time search range for second call becomes $(t_{O_1}, t_{O_1}+T)$. From the historical dataset, let's suppose, we know that the taxi can complete the trip for the first passenger, that is from $O_1$ to $D_1$, in $t(O_1,D_1) = 500s < T$. In this case, it is obvious to assign \emph{take 1} action to the taxi rather than \emph{take 2}. Therefore, we definitely need a dynamic pickup time search range for selecting the second taxi call. After reducing the pickup time search area for second taxi calls, we further reduce the search area by selecting all the taxi trips where taxi can reach before their pickup time $t_{O_2}$ from its intermediate state $\mathbf{s}_{O_1}$. Finally, a second taxi call with the minimum \emph{total extra travel time}, described in following section, is assigned to a taxi.

\begin{figure}[t!]
    \centering
    \begin{subfigure}[t]{0.5\columnwidth}
        \centering
        \includegraphics[width = \columnwidth]{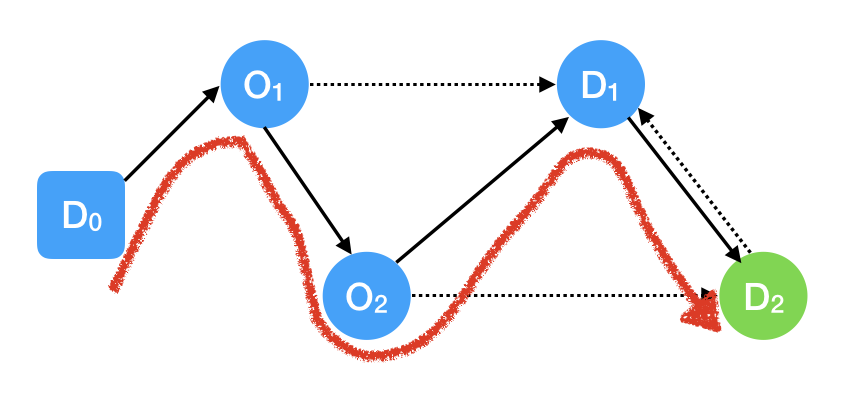}
        \caption{ \emph{Path I}: $O_1$ is dropped first}
        \label{fig:traj1}
    \end{subfigure}%
    ~
    \begin{subfigure}[t]{0.5\columnwidth}
        \centering
        \includegraphics[width = \columnwidth]{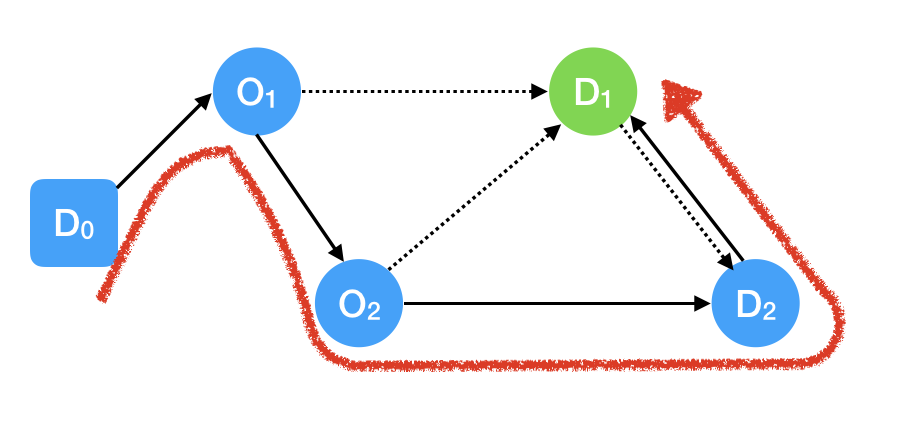}
        \caption{\emph{Path II}: $O_2$ is dropped first }
        \label{fig:traj2}
    \end{subfigure}
    \caption{Possible trajectories}
    \label{fig:traj}
\end{figure}

Now, the taxi has two different passengers on board with different destinations $D_1$ and $D_2$. The next question is which passenger to drop first? This is a routing problem and for simplicity, we consider the solution of this problem is deterministic and embed this decision into the environment, i.e. once an action is assigned to the taxi, the secondary level decision is executed automatically by the environment. The two possible solutions to this routing problem are depicted in Fig. \ref{fig:traj}, that is the taxi can either follow $D_0 \rightarrow O_1 \rightarrow O_2 \rightarrow  D_1 \rightarrow D_2$ (\emph{Path I}) in Fig. \ref{fig:traj1} or $D_0 \rightarrow O_1 \rightarrow O_2 \rightarrow  D_2 \rightarrow D_1$ (\emph{Path II}) in Fig. \ref{fig:traj2}. The final state of the taxi corresponds to the passenger's destination whom is dropped at last, shown in green color for both the solutions.
% for either of two paths is shown in green color. For instance, in \emph{path I}, $D_2$ is the final state of the taxi for the current state transition and it is also the initial state for the next state transition. 

Since the NYC datasets contains trip information only for a selected number of origin and destination pairs, we develop in Section \ref{sec:st-nn} \emph{ST-NN}, a travel time estimation method, which takes raw GPS coordinates of origin and destination and time-of-the-day as input and predicts the travel time.

To choose among these paths, We define the notion of \emph{extra travel time $(\mathrm{Ext_P}(x, y))$} traveled by the taxi going from $x$ to $Y$ when a \emph{path P} is chosen. \emph{Extra travel time $(\mathrm{Ext_P}(\cdot, \cdot))$} is an estimation of extra time each passenger would travel during carpool which otherwise is zero when no carpool. For instance, in Fig. \ref{fig:traj1} the actual travel time for passenger 1, corresponding to $O_1$, is $t(O_1,D_1)$ and for passenger 2, corresponding to $O_2$, is $ t(O_2,D_2)$ when they travel alone. On the other hand, in carpool the travel time for passenger 1, corresponding to $O_1$, is $t(O_1, O_2) + t_{\mathrm{Est}}(O_2,D_1)$ and for passenger 2, corresponding to $O_2$, is  $t_{\mathrm{Est}}(O_2,D_1) + t_{\mathrm{Est}}(D_1,D_2)$. Therefore, the   \emph{extra travel time} for passenger 1 and passenger 2, when \emph{path I} is followed, are 
\begin{align*}
	\mathrm{Ext_I}(O_1, D_1) &= t(O_1, O_2) + t_{\mathrm{Est}}(O_2,D_1) - t(O_1,D_1)\\
	\mathrm{Ext_I}(O_2, D_2) &= t_{\mathrm{Est}}(O_2,D_1) + t_{\mathrm{Est}}(D_1,D_2) - t(O_2,D_2)
\end{align*}
Similarly, when \emph{path II} is followed by the taxi, the extra travel time for both the passengers are given as:
\begin{align*}
	\mathrm{Ext_{II}}(O_1, D_1) &= t(O_1, O_2) + t(O_2,D_2) + t_{\mathrm{Est}}(O_2,D_1) \\ 
	& \qquad \qquad \qquad\qquad \qquad \qquad- t(O_1,D_1)\\
	\mathrm{Ext_{II}}(O_2, D_2) &= t(O_2,D_2) - t(O_2,D_2) = 0
\end{align*}

Now, calculating the individual extra travel time for each of the on-board passengers for both the paths we calculate the total extra travel time as, for \emph{path I} \linebreak $\mathrm{Total_{Ext_I}} =  \mathrm{Ext_I}(O_1, D_1) + \mathrm{Ext_I}(O_2, D_2)$ and for \emph{path II} $\mathrm{Total_{Ext_{II}}} = \mathrm{Ext_{II}}(O_1, D_1) + \mathrm{Ext_{II}}(O_2, D_2)$. Thus, \emph{path I} is followed by the driver if $\mathrm{Total_{Ext_I}} < \mathrm{Total_{Ext_{II}}}$ otherwise \emph{path II} is followed. When \emph{take 1} action is assigned, extra travel time is always zero.

%!TEX root= fullpaper.tex

\section{ST-NN}
\label{sec:st-nn}
We have seen from Section \ref{sec:testbed} that travel time estimation between two points is a key quantity to ensure accurate simulation in the carpool training environment.  In this section, we describe our approach based on deep neural networks for learning travel time for origin-destination pairs that are not part of the NYC dataset.

Deep neural networks are known for solving very difficult computational tasks like object recognition \cite{cirecsan2012deep,jindal2016learning}, regression \cite{west2000neural} and other predictive modeling tasks. They do so, because of their high ability to learn feature representations from the data and best map the input features to the output variables.

\begin{figure}[t!]
\centering
\includegraphics[width=\columnwidth]{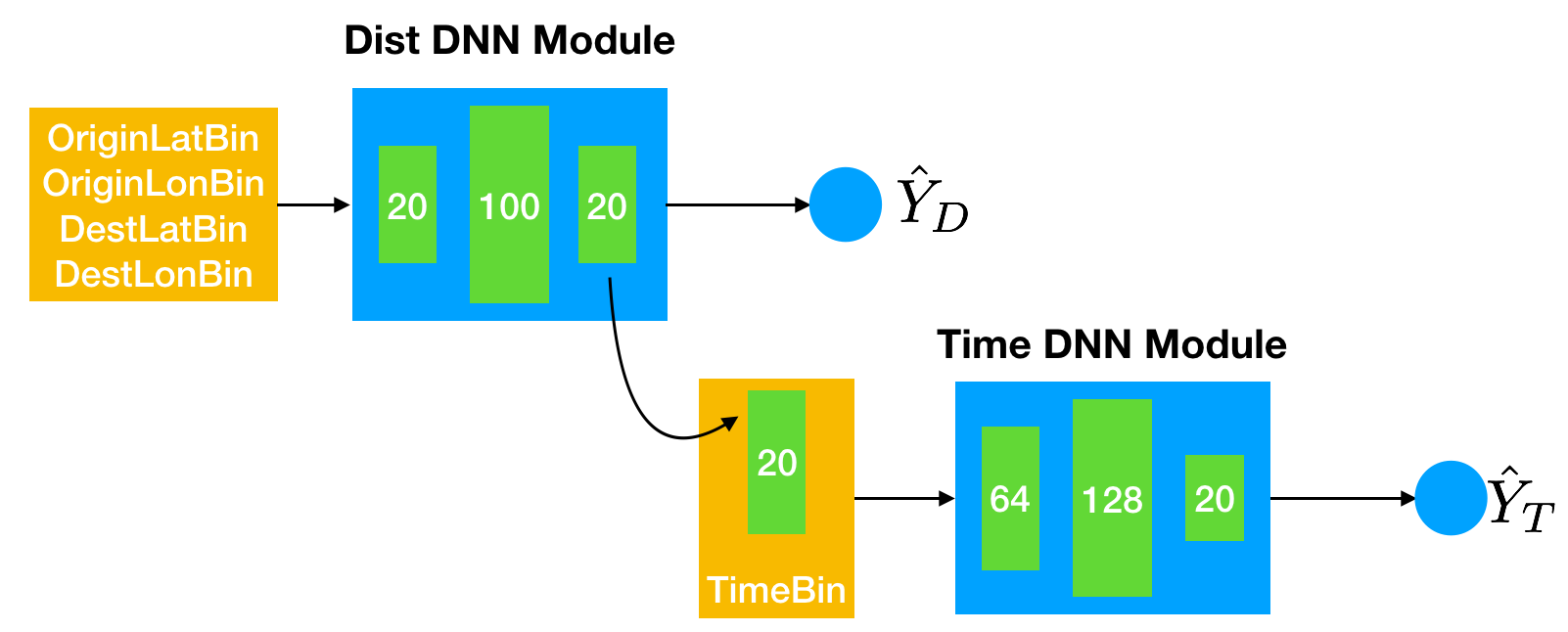}
\caption{Unified neural network architecture for joint estimation of travel time and distance}
\label{fig::ProbSchem}
\vspace{-0.1in}
\end{figure}

In Fig. \ref{fig::ProbSchem}, we describe the ST-NN architecture. In this architecture, we define two different deep neural network (DNN) module both for travel distance and travel time estimation as ``Dist-DNN Module" and ``Time-DNN Module'', respectively. First, we describe the input to both the two modules. The input to dist-DNN module is only the origin $o_i$ and destination $d_i$ binned GPS coordinates. This module is not exposed to time-of-day $t_i$ information because the time-of-day information is irrelevant to the travel distance estimation and might misguide the network. Any taxi service platform, because of usual reasons, always routes a driver on to a path of shortest length. As the route planning is not a part of this work, we assume that the all the taxis in the available taxi trip dataset have chosen the shortest path for a trip irrespective of the time-of-day. Therefore, the input dimension to dist-DNN module is a 4-D vector, that is OriginLatBin, OriginLonBin, DestLatBin and DestLonBin. The input to time-DNN module is the activations of last hidden layer of the dist-DNN module, which encodes the raw GPS coordinates into a feature vector, concatenated with the time-of-day information. 

Here, both the dist-DNN module and time-DNN module are three-layer MLPs with different numbers of neurons per layer. We cross-validated the parameters and found the ones with the best performance. The best performance configuration of the number of layers and number of neurons per layer for both the module is shown in Fig. \ref{fig::ProbSchem} where, $\hat{Y}_D$ and $\hat{Y}_T$ are the predicted distance and time from dist-DNN module and time-DNN module, respectively. The ST-NN architecture is then trained via stochastic gradient descent jointly for both travel distance and time according to the loss function:
\begin{align*}
&L(Y_D, Y_T,\hat{Y}_D,\hat{Y}_T)= L(Y_T, \hat{Y}_T) + L(Y_D ,\hat{Y}_D)\\
% \end{equation}
% From \eqref{eq:loss}, we write the final loss function as:
% \begin{align}
& = \frac{1}{2N}\sum_{i=1}^N (Y_T^i-\hat{Y}_T^i)^2 + \frac{1}{2N}\sum_{i=1}^N (Y_D^i-\hat{Y}_D^i)^2.
\end{align*}	
% \begin{multline}
% & = \frac{1}{2N}\sum_{i=1}^N (Y_T^i-\hat{Y}_T^i)^2 + \frac{1}{2N}\sum_{i=1}^N (Y_D^i-\hat{Y}_D^i)^2
% \end{align}	

We observe, in Section \ref{sec::st-nn_results}, that the joint learning of travel time and distance as described in Fig. \ref{fig::ProbSchem} improves the travel time estimation over the baseline methods.

\section{Q-Learning for Carpooling} 
\label{sec:q-learning}
In this work, we consider that the taxi is completely relying on RL in order to decide on carpooling by learning the value function of a taxi's state-action pair from the gathered experience generated from the carpooling simulator. 
% In reinforcement learning, an agent observes a state $\mathbf{S}_t$, takes an action $a$, receives a reward $r$ and move to next state $\mathbf{S}_{t+1}$. 
We adopt a model-free RL approach to learn an optimal policy as the agent has no knowledge about the state transitions and reward distributions. A policy $\pi$ is a map which models the agent's action selection given a state where the value of a policy is determined by the state-action value function $V^\pi(\mathbf{s}) = E[R|\mathbf{s};\pi]$. Here, $R$ denotes the sum of discounted reward. The value function estimates how good for an agent to be in a given state following the policy $\pi$. Given an optimal policy $\pi^*$ and an action in a given state $\mathbf{s}$, the action-value under an optimal policy is defined by $Q^*(\mathbf{s},a) = E[R|\mathbf{s};a,\pi^*]$.  The optimal action can be found by $\arg\max_a Q^*(\mathbf{s},a)$.  With tabular Q-learning, where the Q-value function $Q(\mathbf{s},a)$ is estimated by updating the lookup table as 
\begin{equation}
	Q(\mathbf{s}_t,a) := Q(\mathbf{s}_t,a) + \alpha[r+\gamma\max_aQ(\mathbf{s}_{t+1},a)- Q(\mathbf{s}_t,a)].
\end{equation}   
Here, $0\leq \gamma< 1$ is the discount rate, modeling the behavior of the agent when to prefer long term reward $(\gamma \rightarrow 1)$ than immediate reward $(\gamma = 0)$ and $0< \alpha \leq 1$ is the step size parameter which controls the learning rate. In training, we use the epsilon-greedy policy, where with probability $1-\epsilon$, an agent in state $s$ selects an action $a$ having the highest value $Q(\mathbf{s},a)$ (\emph{exploitation}), and with probability $\epsilon$ choose a random action to ensure \emph{exploration}. 

Tabular Q-learning is good for small MPD problems but with the huge state-action space or when the state space is continuous we use a function approximator to model the $Q(\mathbf{s},a) = f_{\theta}(\mathbf{s},a)$.
 % rather than learning a lookup table, here $\theta$ denotes the parameters of function approximator. 
The best example of function approximator is neural networks (universal function approximator). Here, we adopt the basic neural network architecture in \cite{mnih2015human}, where the neural network takes the state space (longitude, latitude, time of day) as input and output multiple Q values corresponding to the actions $(W, TK1, TK2)$. To approximate the Q function we use a three-layer deep neural network which learns the state-action value function. As in \cite{mnih2015human}, we stored the state transitions (experiences) in a replay memory and each iteration samples a mini-batch from this replay memory. In the DQN framework, the mini-batch update through back-propagation is essentially a step for solving a bootstrapped regression problem with the loss function 
\begin{equation} 
 	(Q(\mathbf{S}_t,a|\theta) - r(\mathbf{s}_t,a) - \gamma \max_aQ(\mathbf{S}_{t+1},a|\theta^\prime))^2, 
\end{equation}
where $\theta^\prime$ is the parameters for the Q-network of the previous iteration. 

Here the max operator is used both for selecting and evaluating an action which makes the Q-network training unstable. To improve the training stability we use Double-DQN as proposed in \cite{van2016deep} where a target $Q$-network $\hat{Q}$ is maintained and synchronized periodically with the original $Q$-network. Thus the modified mini-batch target is
\begin{equation}
	r(\mathbf{s}_t,a) + \gamma \hat{Q}(\mathbf{S}_{t+1},\arg \max_a Q(\mathbf{s}_{t+1},a|\theta^\prime) | \hat{\theta}^\prime).
\end{equation}
% DQN is explained in Algorithm \ref{DQN}. As we want 
To maximize total effective trip distance, we set the discount factor $\gamma = 0.95$ for all the experiments. We summarize the DQN algorithm in Algorithm \ref{DQN}. We compare the performance of DQN learned policy with respect to a fixed policy which always favors carpooling. Details of fixed policy generation is described in Algorithm \ref{Fixed_Policy}.

\begin{algorithm}
\caption{DQN}\label{DQN}
\begin{algorithmic}[1]
\Procedure{}{}
\BState \textit{Initialize}: Replay Memory $\mathcal{M}$, $Q$ function with random weights $\theta$, target $\hat{Q}$ function with weights $\hat{\theta} = \theta$
\BState \textit{Episode}:
\State $\textit{$l_0$} \gets  \textit{$random$} $, $\textit{$t_0$} \gets  \textit{$0$} $, $\textit{$\mathbf{S}_0$} \gets  \textit{$(l_0,t_0)$ } $
\BState \textit{loop}: $t = 1:T$
\BState \textit{Follow}: \emph{$\epsilon$-greedy}
\State  \[a_t \gets \begin{cases}
        \text{random action}, & \text{ probability $\epsilon$}\\
        \arg \max_a Q(\mathbf{s}_t,a;\theta), & \text{otherwise}
        \end{cases} \]
\State \textit{$\mathbf{s}_{t+1}, r_t$} $\gets$ \text{Simulator}($\mathbf{s}_t,a_t$) 
\State \textit{$\mathcal{M}$} $\gets$ \textit{$\{\mathbf{s}_t,a_t,r_t,\mathbf{s}_{t+1}\}$}  \Comment Replay Buffer
\State \text{Select minibatch of state transitions} $\{\mathbf{s}_t,a_t,r_t,\mathbf{s}_{t+1}\} \in \mathcal{M}$
\BState \textit{Form a target $y_i$}
\State  \[y_i \gets \begin{cases}
        r_i, & \text{ if episode ends}\\
        r_i+ \gamma\max_aQ(\mathbf{s}_{i+1},a;\hat{\theta}), & \text{otherwise}
        \end{cases} \]
\State \textit{Loss($\theta$)} $\gets$ $(y_i - Q(\mathbf{s}_i,a_i;\theta))^2$
\State $\hat{\theta} \gets \theta$ \Comment Periodic update of network
\EndProcedure
\end{algorithmic}
\end{algorithm}

\begin{algorithm}
\caption{Fixed Policy}\label{Fixed_Policy}
\begin{algorithmic}[1]
\Procedure{}{}
\BState \textit{Given}: $\mathcal{D}$, $l_0$, $t_0$, $t_d$, $T = 600$ Sec., $T_c$
\BState \textit{Episode}:
\State $\textit{$l_0$} \gets  \textit{$random$} $, $\textit{$t_0$} \gets  \textit{$0$} $, $\textit{$\mathbf{S}_0$} \gets  \textit{$(l_0,t_0)$ } $
\BState \textit{loop}:
\If {\textit{\emph{Take 1} is not possible}}
    \State \textit{Do wait action} \Comment Algorithm \ref{W_Action}
\Else 
    \If {\textit{\emph{Take 2} action is possible}}
        \State \textit{Do Take 2 action} \Comment Algorithm \ref{TK1_Action}
    \Else 
        \State \textit{Do Take 1 action} \Comment Algorithm \ref{TK2_Action}
    \EndIf
\EndIf
\State $\textit{$\mathbf{S}_0$} \gets  \textit{$\mathbf{S}_1$}$
\If {$(t_0 \leq 24*60*60)$}
    \State \textbf{goto} \emph{loop}.
\Else
    \State \textbf{goto} \emph{Episode}.
\EndIf
\EndProcedure
\end{algorithmic}
\end{algorithm}
%!Tex root = fullpaper.tex

\section{Performance Evaluation}
\label{sec:performance}
Based on the NYC dataset, we have conducted experiments first on travel time estimation using ST-NN and then on first-level carpool policy optimization using reinforcement learning. We report the detailed results below.

\subsection{ST-NN Results} % (fold)
\label{sec::st-nn_results}
We divide the entire dataset into training and test subsets in the ratio 80:20. All the parameters of ST-NN network architecture such as the number of layers per module and the number of units per hidden layer are shown in the Fig. \ref{fig::ProbSchem}. We cross-validated the hyper-parameters to achieve the best performance. We also use data mapping as described in Section \ref{subsetion:mapping}. For location mapping, we use $(200m \times 200m)$ 2-D square cell and for time mapping we use 10 minutes as 1-D time cell. All the parameters of ST-NN are kept fixed throughout all the experiments. 

% subsection evaluation_methods (end)
\subsubsection{Outliers Rejection}
\label{sec:::outlier}
From the initial exploration of NYC taxi trip data we find that the dataset contains a number of anomalous taxi trips termed as outliers, for example having more than 7 passengers in a taxi and no passenger, missing pickup and drop-off GPS coordinates, travel time of zero seconds while the corresponding travel distance is non-zero, travel distance of zero miles while corresponding travel time is non-zero. These outliers can cause huge mistakes in our estimations, so we experimentally detected the anomalous trips and removed them from the dataset.

\begin{table}[!h]
\begin{tabular}{|c|c|c|c|c|c|c|}
\hline
                                     \multicolumn{2}{|c|}{}       & \begin{tabular}[c]{@{}c@{}}$R^2$\\  Coef.\end{tabular} & MAE    & MRE   & MedAE   & MedRE  \\ \hline
\multirow{3}{*}{Time}     & LRT    & -1.84                                                    & 724.14 & 1.01  & 638.52  & 1.10   \\ \cline{2-7} 
                                     & TimeNN & 0.71                                                    & 158.29 & 0.22 & 100.24 & 0.18  \\ \cline{2-7} 
                                     & ST-NN  & \textbf{0.75}                                                     & \textbf{145.9}  & \textbf{0.20}  & \textbf{91.48}   & \textbf{0.16 }  \\ \hline %\hline

\end{tabular}
 \caption{Overall performance comparison of ST-NN with the other approaches for travel time estimation.}
     \label{tabel:TimeDist_comp}
\end{table}
\subsubsection{Evaluation Methods} % (fold)
\label{sec:::evaluation_methods}
Here we list the methods compared to ST-NN:
\begin{enumerate}
    \item[a.] Linear Regression for Time (\emph{LRT}): We implement a simple linear regression method for time estimation. 
    \item[b.] Unified learning (\emph{ST-NN}): This is the proposed approach described in Section \ref{sec:st-nn}.
    \item[c.] Time-DNN module (\emph{TimeNN}): When only the time-DNN module of the ST-NN is used to learn the travel time. Inputs to this module are the origin and destination GPS coordinates along with time-of-day.
    \item[d.] BTE : We also compare the performance of ST-NN with the best method introduced in \cite{wang2016simple}.
\end{enumerate}

\subsubsection{Evaluation Measures}
\label{sec:::Performance_evaluation}

We evaluate the performance of ST-NN on five different metrics, \emph{Mean Absolute Error (MAE)} defined as the mean of the absolute difference between the estimated travel time $f_i$ and the ground truth $y_i$, 
\begin{equation*}
    \mathrm{MAE} = \frac{\sum_{i=1}^N |y_i - f_i|}{N}
\end{equation*}
and, \emph{Mean Relative Error (MRE)} is defined as:
\begin{equation*}
    \mathrm{MRE} = \frac{\sum_{i=1}^N |y_i - f_i|}{\sum_{i=1}^N y_i}
\end{equation*}
Since the dataset contains anomalous taxi trip entries we also measure \emph{Median Absolute Error (MedAE)} and \emph{Median Relative Error (MedRE)} as $$ \mathrm{MedAE} = \mathrm{median}( |y_i - f_i|),$$ $$\mathrm{MedRE} = \mathrm{median}\left(\frac{|y_i - f_i|}{y_i}\right),$$ where $\mathrm{median} $ has its usual meaning. Finally, to measure how close the data are to the fitted hyper surface, we also use the coefficient of determination $R^2$ to evaluate the performance of ST-NN: $ R^2 = 1- \frac{\sum_{i}(y_i - f_i)^2}{\sum_{i} (y_i - \bar{y})^2},$ where $\bar{y} = \frac{1}{N}\sum_{i=1}^N y_i$, is the mean of the observed data. 

Table \ref{tabel:TimeDist_comp} compares the performance of proposed approach for travel time estimation. From Table \ref{tabel:TimeDist_comp}, we observe that TimeNN is far better than the simple linear regression method for travel time estimation, e.g. about $78\%$ improvement in MAE. This is expected because the simple linear regression does not consider the uncertain traffic conditions and simply tries to find the linear relationship between the raw origin-destination GPS coordinates and the travel time.  

\begin{figure}[]
    \centering
    \begin{subfigure}[t]{0.45\columnwidth}
        \centering
        \includegraphics[width = \columnwidth]{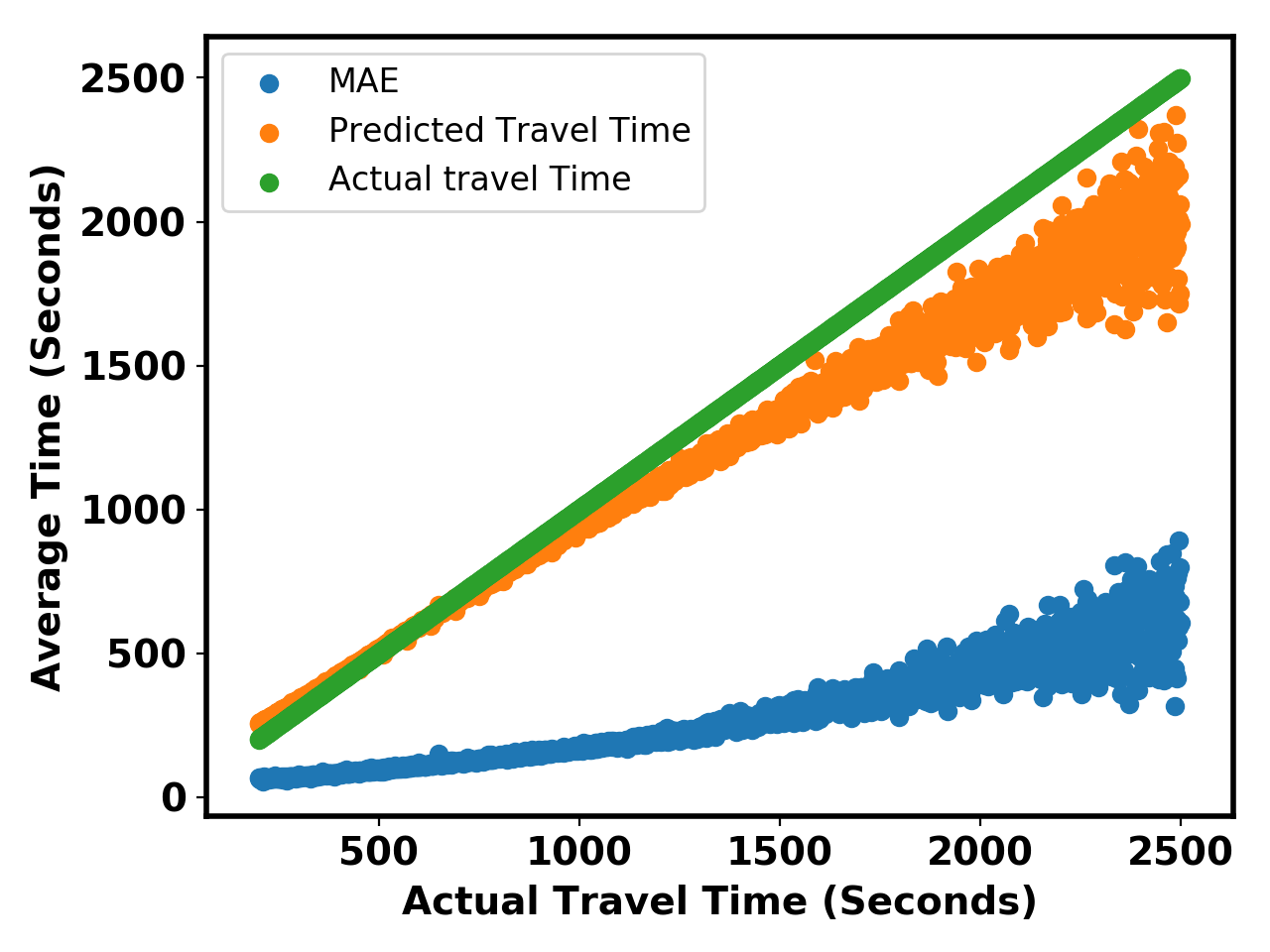}
        \caption{ }
        \label{fig::TraveMAE}
    \end{subfigure}
    ~ 
    \begin{subfigure}[t]{0.45\columnwidth}
        \centering
        \includegraphics[width = \columnwidth]{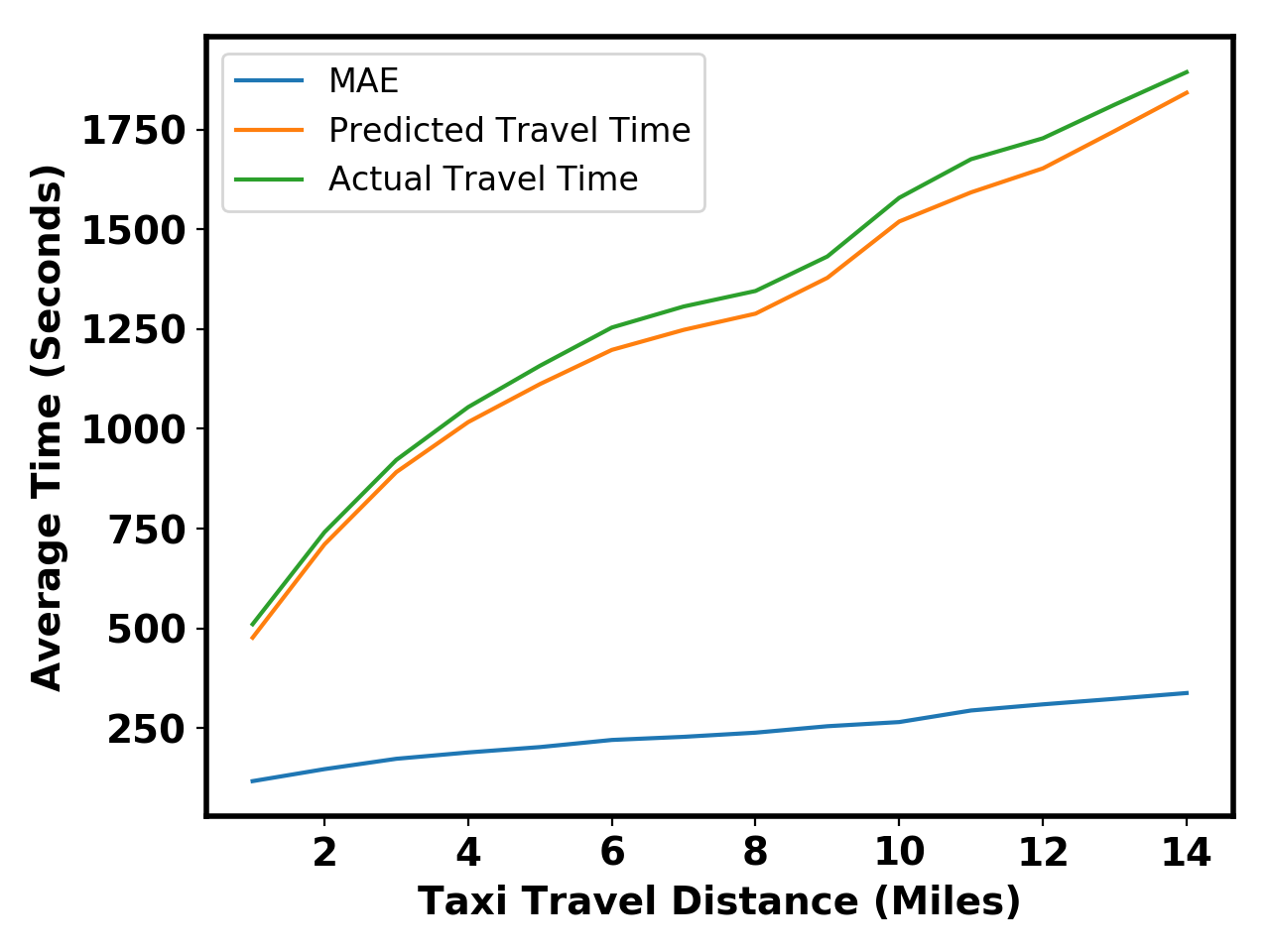}
        \caption{}
        \label{fig::DistMAE}
    \end{subfigure}%
    
  \begin{subfigure}[t]{0.5\columnwidth}
      \centering
      \includegraphics[width = \columnwidth]{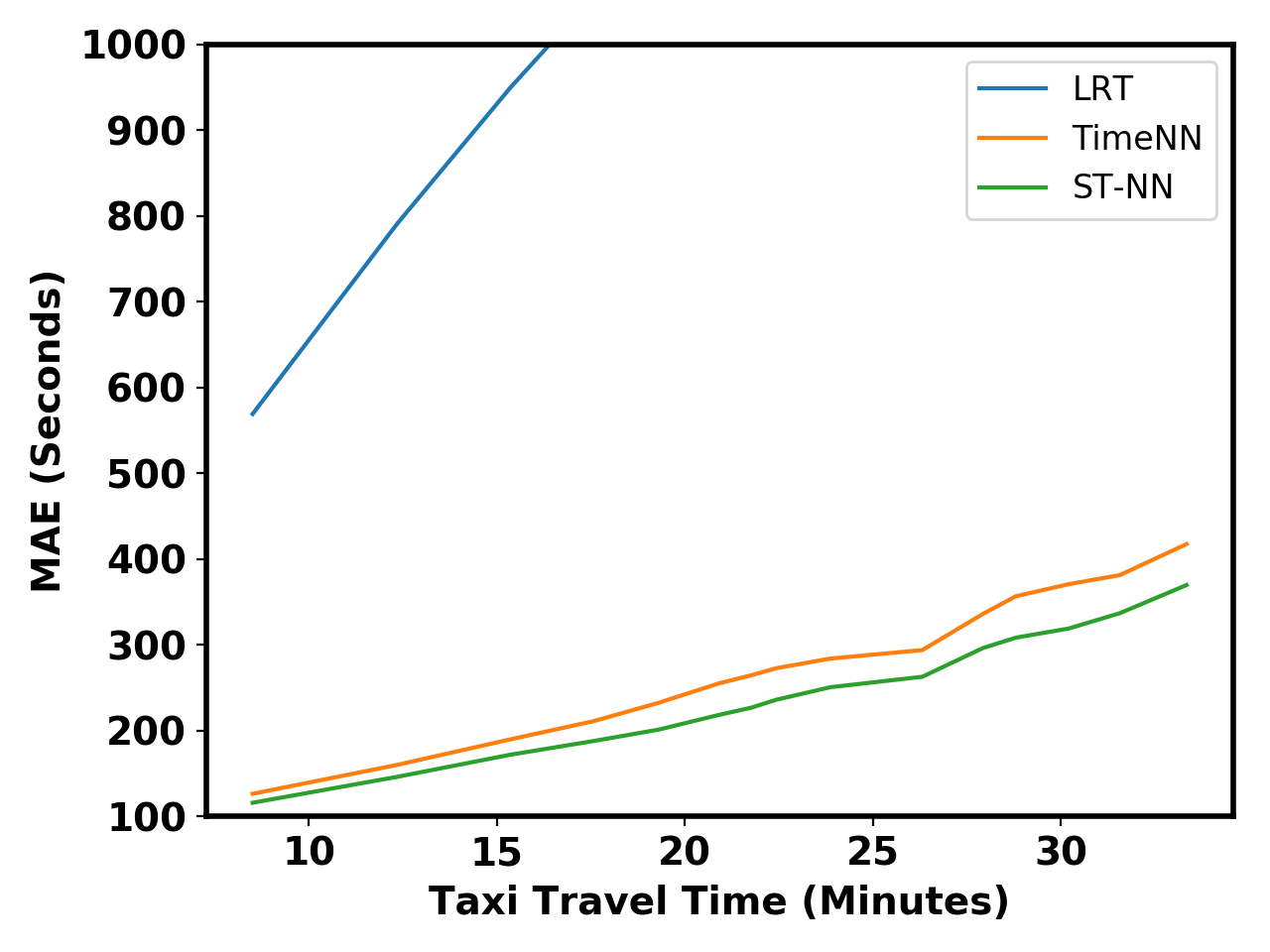}
      \caption{}
      \label{fig::TimeComp}
  \end{subfigure}%
    \caption{ST-NN performance as a function of (a) Travel time; (b) Travel distance (c) Overall comparison}
    \label{}
\end{figure}

With encoded travel distance information, ST-NN further improves the performance for travel time estimation, that is MAE is improved by 13 seconds in comparison to TimeNN. To investigate further, we plot the MAE for all the approaches in Fig. \ref{fig::TimeComp} to know in which regimes the ST-NN is better than the TimeNN. It is clear from the plots that the slope of the orange curve is larger than the green curve, which means that the longer a trip lasts, the more significant gap in performance is noticed. We also plot the MAE and predicted travel time for ST-NN network as a function of taxi travel time in Fig. \ref{fig::TraveMAE}. As expected, for the shorter taxi trips, ST-NN succeeds in predicting the actual travel time but for the longer travel trips, it encounters a larger MAE, around $8-10$ minutes. In Fig. \ref{fig::DistMAE}, we show the performance of ST-NN for travel time estimation with respect to the trip distance. We obtain similar observation in performance.

\begin{table}[h]
    \centering
    \begin{tabular}{|c|c|c|c|c|c|}
         \hline
        \multicolumn{2}{|c|}{}                                                                    & MAE    & MRE &MedAE&MedRE    \\ \hline
        \multirow{2}{*}{\begin{tabular}[c]{@{}c@{}}With \\ Outliers\end{tabular}}   & BTE \cite{wang2014travel} & 170.04 & 0.2547 &97.435 &0.196\\ \cline{2-6} 
                                                                                    & ST-NN & \textbf{123.13} & \textbf{0.2282} & \textbf{81.21} & \textbf{0.183}\\ \hline
        \multirow{2}{*}{\begin{tabular}[c]{@{}c@{}}Without\\ outliers\end{tabular}} & BTE \cite{wang2014travel}  & 142.73 & 0.2173 & 90.046&0.1874\\  \cline{2-6} 
                                                                                    & ST-NN   & \textbf{121.48} & \textbf{0.2155} &\textbf{80.77} &\textbf{0.182} \\ \hline

    \end{tabular}
    \caption{ST-NN Vs. BTE \cite{wang2014travel} }
    \label{tabel:w_o_comp}
\end{table}

We also compare the performance of ST-NN with the best approach in \cite{wang2014travel} and study the impact of outliers on the performance of ST-NN for travel time estimation in Table \ref{tabel:w_o_comp}. For a fair comparison we mask the training dataset confined only to Manhattan region and use the same data mapping parameters as described in \cite{wang2014travel}. In Section \ref{sec:::outlier}, we studied the types of outliers present in dataset and applied certain filters on the dataset such as filters using time and distance, GPS coordinates etc to remove the outliers. To analyze the robustness of ST-NN with respect to outliers, we train the ST-NN on the cleaned training data and test the network on uncleaned (with outliers) data. Without outliers, we observe a clear performance improvement of ST-NN for travel time estimation, in terms of MAE, by 17\%. We found that even when the outliers are prevalent in the data, our proposed approach not only outperformed \cite{wang2014travel} but also appeared to be more robust to outliers. We observe a negligible difference in the performance of ST-NN with or without outliers ($\sim 2$ seconds in MAE).

\subsection{Carpooling Results}
With ST-NN, we deployed the trip time estimation module to our carpool training environment developed in Section \ref{sec:testbed}. We trained an RL agent using experiences generated from the simulation environment to optimize the carpooling policy of a single taxi driver. In this work, we consider a single agent carpooling policy search where the decision taken by an agent (taxi) is independent of the other agents. In a single agent or multi-agent RL learning framework agent is a ride-sharing platform which takes decision for the taxis. In our problem when ride-sharing platform takes decision for only a single taxi then taxi itself acts as an agent. 
For learning a tabular-Q policy, we discretized the selected geographical region into square cells of 0.0002 degree latitude $\times$ 0.0002 degree longitude (about 200 mt.$\times$ 200 mt.) forming a 2-D grid and also discretized the time of day with 600s as sampling period, whereas for learning a DQN policy we do not discretize any of the variables. The original variable values are used as input to the agent neural network.

\begin{figure}[h]
    \centering
    \begin{subfigure}[]{0.49\columnwidth}
        \centering
        \includegraphics[width = \columnwidth]{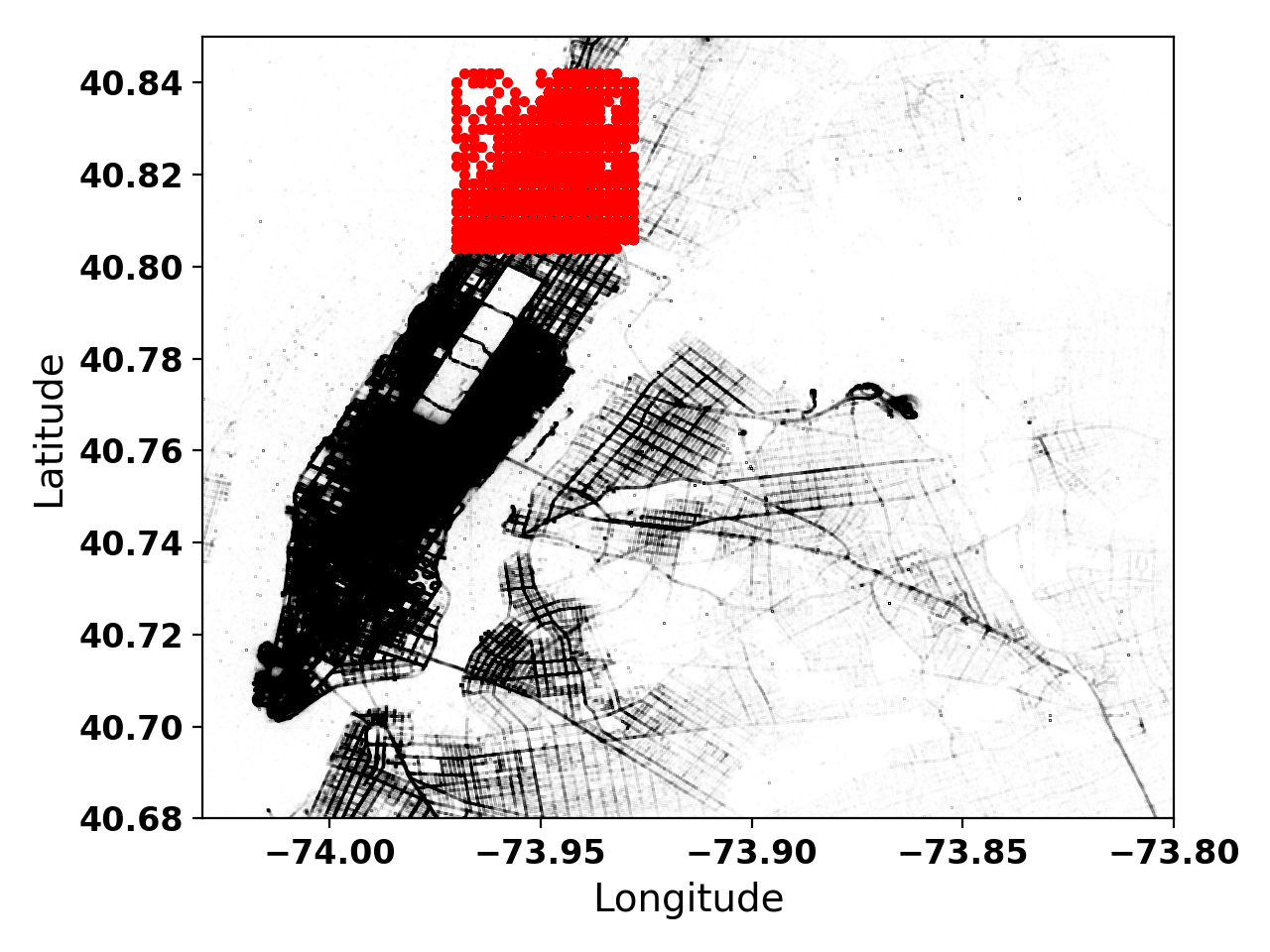}
        \caption{Uptown Manhattan}
        \label{fig:North}
    \end{subfigure}%
    ~ 
    \begin{subfigure}[]{0.49\columnwidth}
        \centering
        \includegraphics[width = \columnwidth]{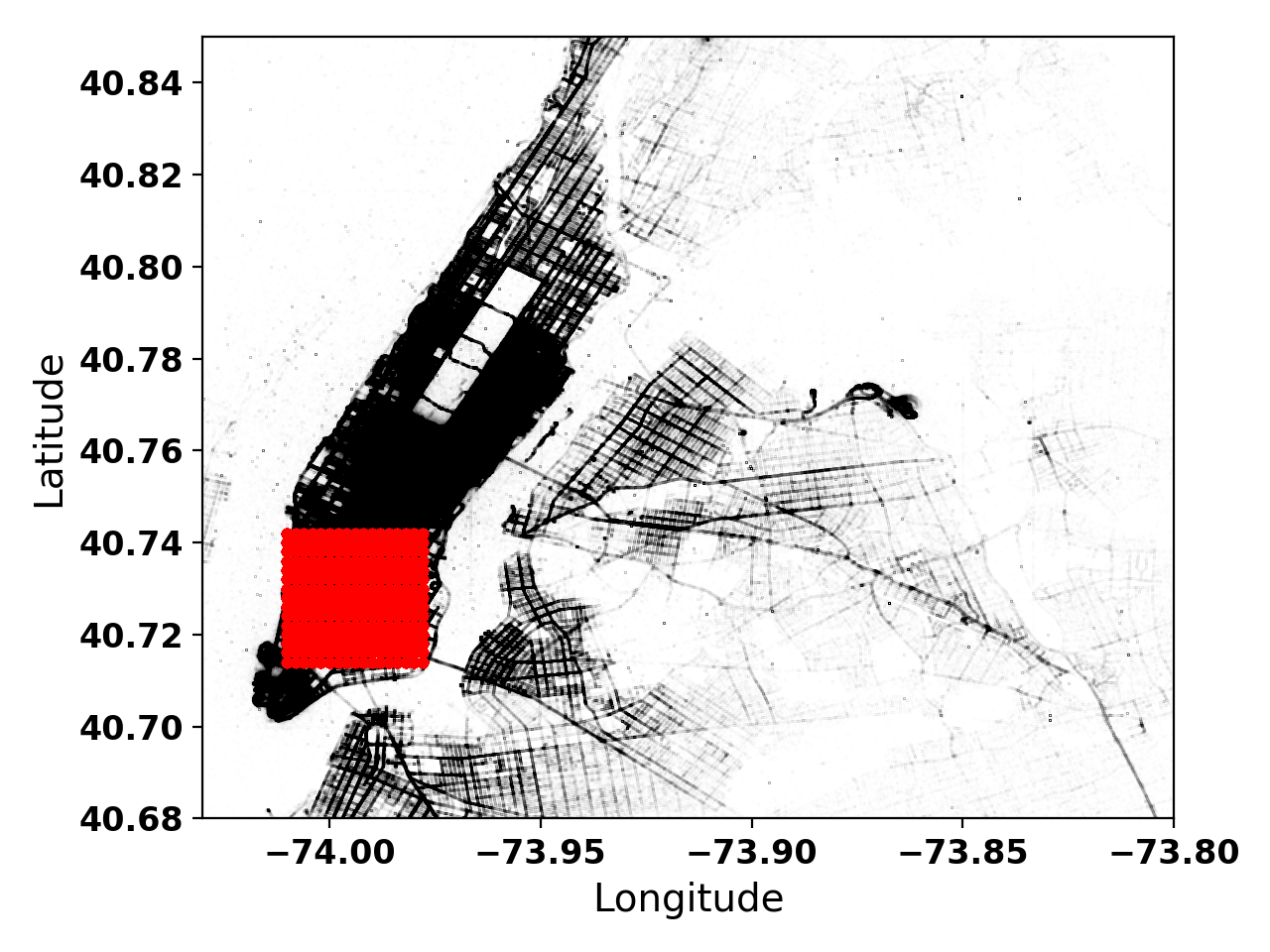}
        \caption{Downtown Manhattan}
        \label{fig:Down}
    \end{subfigure}
    \caption{Red dots represented the selected region. [Best seen in color]}
    \label{fig:NorthDown}

\end{figure}

We evaluate the performance of DQN learned policy both on weekday and weekend by comparing the mean cumulative reward with respect to the fixed policy (baseline) that always favors carpooling and the tabular-Q policy. By far, fixed policy is the greedy policy in the sense that the agent always chooses an action which always accepts a carpool (associated with the maximum immediate reward) as described in Algorithm \ref{Fixed_Policy}.

We generate the samples of experience in real-time from the carpool simulation environment described in Section \ref{sec:testbed}. We study the performance of learned RL policy for two different taxi call densities regions in NYC, Uptown Manhattan and Downtown Manhattan in Fig. \ref{fig:NorthDown}.

\subsubsection{Uptown Manhattan}
We select a square region in northern Manhattan in longitude $[-73.9694, -73.9274]$ and in latitude $[40.805, 40.8438]$ as shown in Fig. \ref{fig:North} where binned red dots represent the selected region (about $20 \times 20$ grid). 

\begin{figure}[h]
    \centering
    \begin{subfigure}[]{0.5\columnwidth}
        \centering
        \includegraphics[width = \columnwidth]{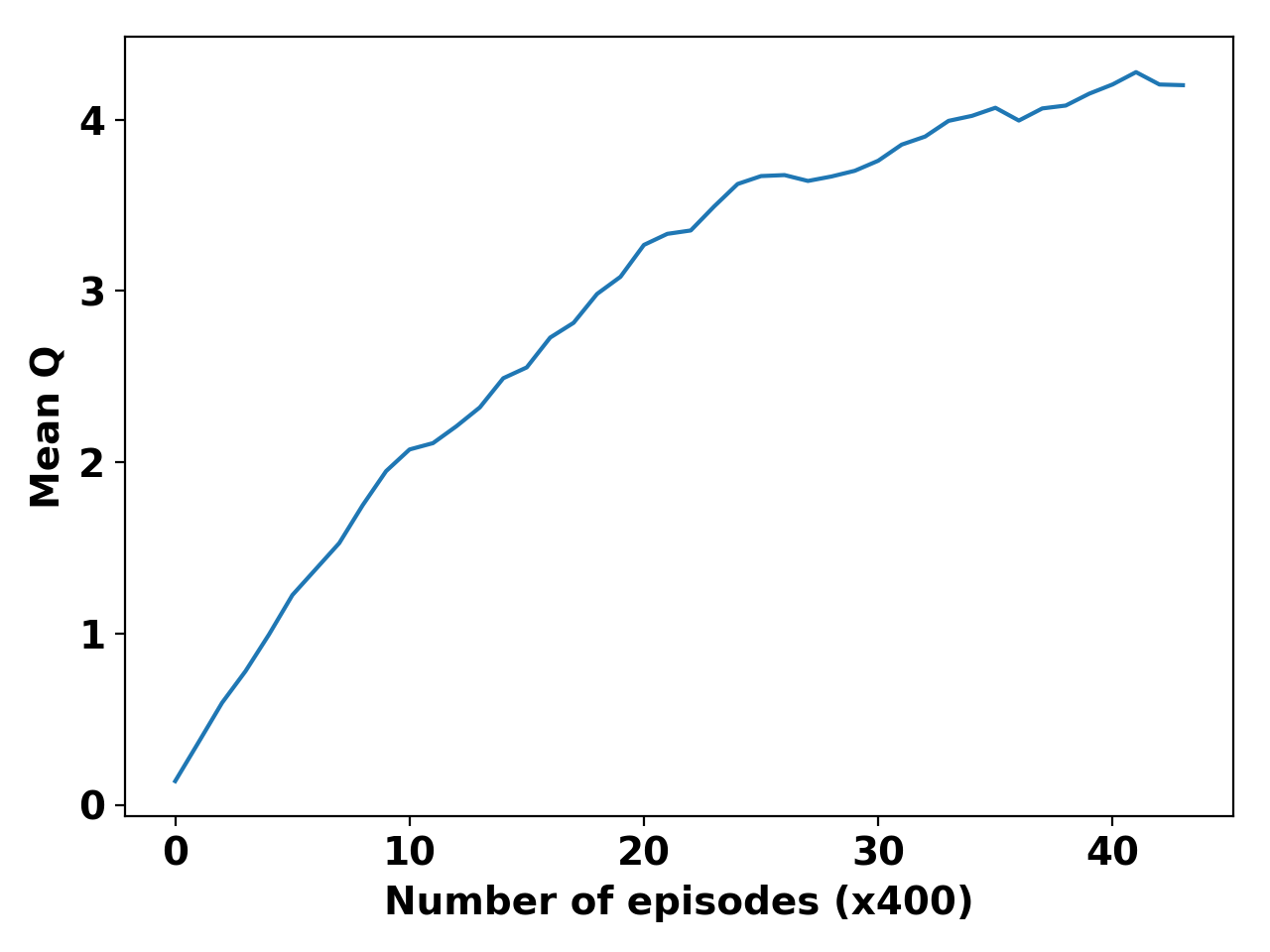}
        \caption{DQN}
        \label{fig:N_meanDQ}
    \end{subfigure}%
    ~ 
    \begin{subfigure}[]{0.5\columnwidth}
        \centering
        \includegraphics[width = \columnwidth]{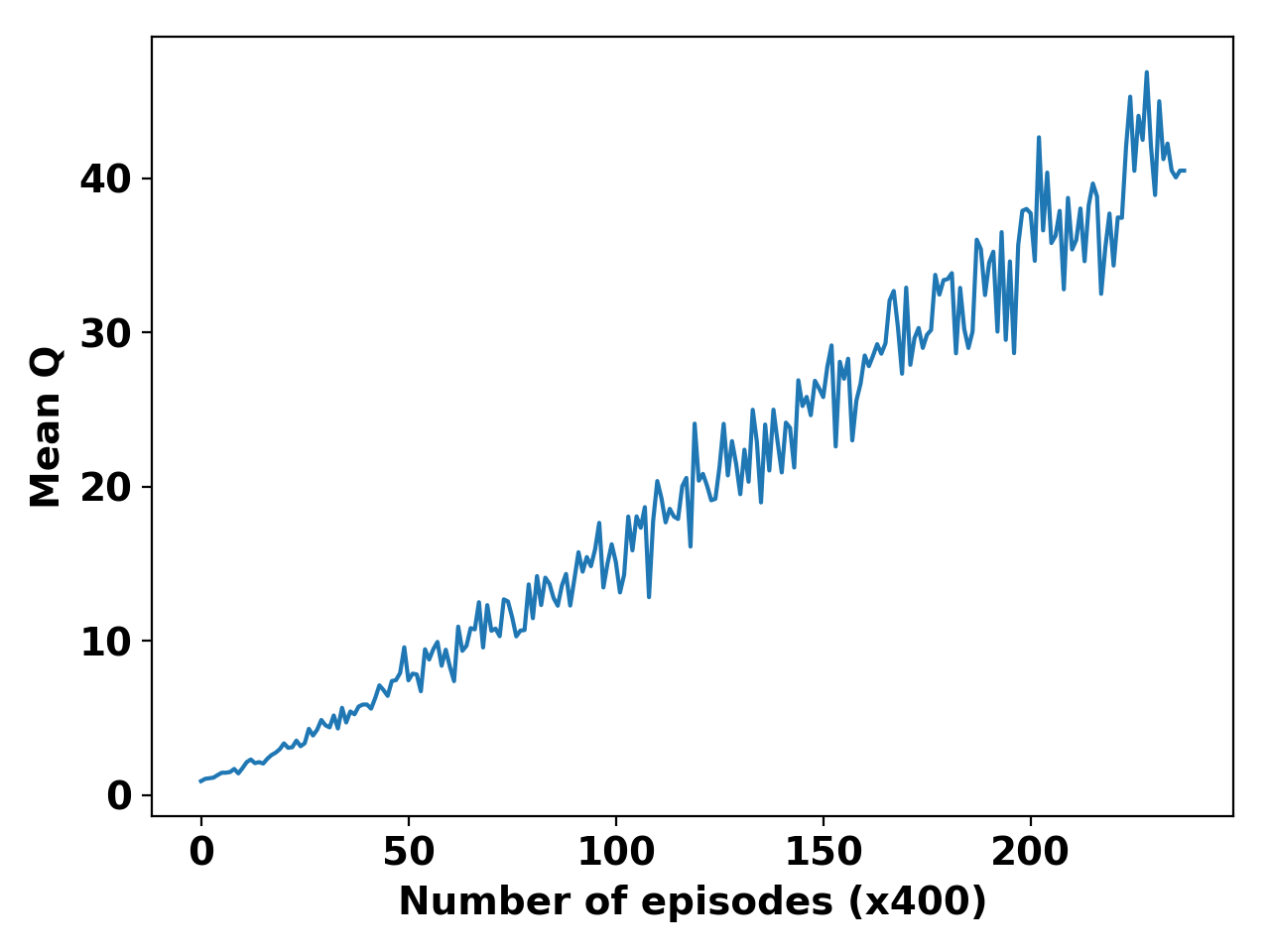}
        \caption{Tabular Q}
        \label{fig:N_meanTQ}
    \end{subfigure}%
%     \caption{Uptown Manhattan: Mean Q}
%     \label{fig:N_DQN}
    
    \begin{subfigure}[]{0.5\columnwidth}
        \centering
        \includegraphics[width = \columnwidth]{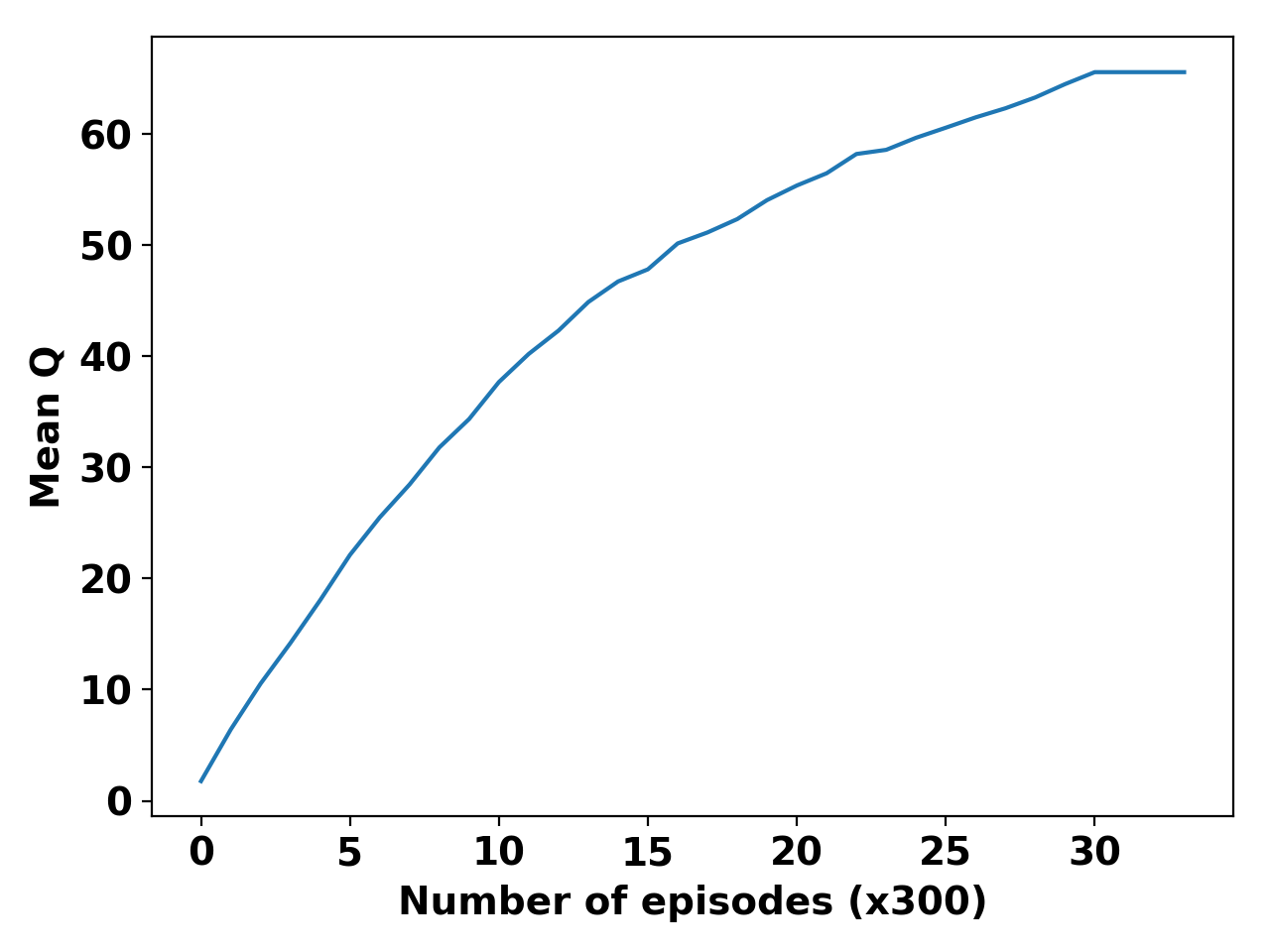}
        \caption{DQN}
        \label{fig:D_MeanDQ}
    \end{subfigure}%
    ~ 
    \begin{subfigure}[]{0.5\columnwidth}
        \centering
        \includegraphics[width = \columnwidth]{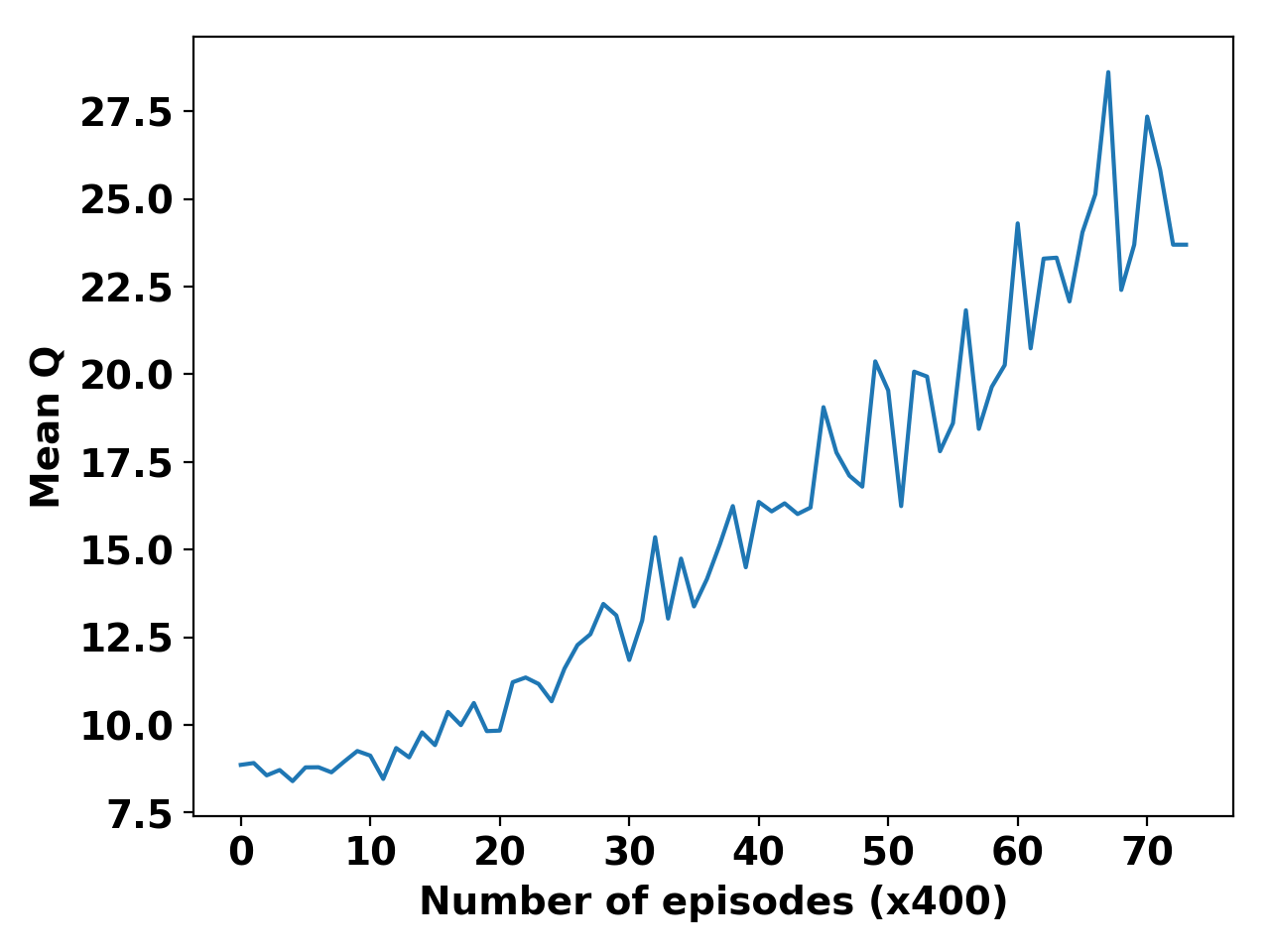}
        \caption{Tabular Q}
        \label{fig:D_MeanTQ}
    \end{subfigure}%
    \caption{Mean Q: (Row I) Uptown and (Row II) Downtown Manhattan}
    \label{fig:D_DQN}

\end{figure}

In table \ref{table:NorthReward}, the first row compares the performance of DQN learned policy both for weekday and weekend with respect to the fixed policy. The DQN learned policy outperforms both the fixed policy and the tabular Q policy both on weekday and weekend. We plot the action-values (Q-value) averaged over mini-batches for DQN in Fig. \ref{fig:N_meanDQ} and for the Tabular Q, Q-value is averaged over a number of episodes in Fig. \ref{fig:N_meanTQ} for a weekday. In both the cases, mean Q smoothly converged after few thousand episodes and we stop the training of the RL agent. 

\subsubsection{Downtown Manhattan} We select a square region of Downtown Manhattan in longitude $~[-74.0094, ~-73.9774~]$ and in latitude $[40.715, 40.7438]$ as shown in Fig. \ref{fig:Down}. 

Similar to uptown Manhattan, we plot the action-values in Fig. \ref{fig:D_MeanDQ}, \ref{fig:D_MeanTQ} for DQN and Tabular Q on a weekday respectively. In table \ref{table:NorthReward}, second row compares the performance of DQN learned policy both for weekday and weekend with respect to fixed policy and tabular Q learned policy. On weekday DQN and the fixed policy performed equally well.  This can be explained by the fact that downtown Manhattan has a high density of taxi calls, and their destinations are usually within the downtown area as well.  Hence, an always-carpool policy is near-optimal in optimizing the objective, i.e. the effective total trip distance.  On the other hand, during the weekend taxi calls density is reduced, and DQN learned an optimal policy better than the baseline.

\begin{table}%[h!]
    \centering
    \renewcommand{\arraystretch}{}
    \scalebox{1}
    {
    \begin{tabular}{|c|c|c|c|l|}
        \hline
                Region        &        Day           & Fixed Policy & Tabular Q & DQN  \\ \hline
                   Uptown         & Weekday           & 41.543       & 39.17     & \textbf{46.08}  \\ \cline{2-5}
                            & Weekend           &25.39       & 14.37     & \textbf{27.86}   \\
                             \hline %\cmidrule(lr){2-7}

            Downtown        & Weekday        & \textbf{340.06}       & 186.00     & {339.42}   \\ \cline{2-5}
                            & Weekend        & 259.57       & 145.63    & \textbf{261.23}  \\ 
                         \bottomrule
        \end{tabular}
}
\caption{Mean cumulative reward on weekday and weekend}
\label{table:NorthReward}

\end{table}
Tabular-Q performance is always worst because the state-action space is huge and obtaining Q value for such a state-action space is not practical. In all the experiments, we learned a very sparse Q value table. Therefore, at test time we encounter some states where the Q values for all the actions are equal to zero. 

We suspect that in downtown Manhattan where the taxi calls are very frequent, DQN policy always favors for carpool and generate the reward similar to fixed policy. On the other hand, in uptown Manhattan where taxi calls are less frequent, DQN learned policy is able to selectively take $TK1$ or $W$ action, leading the taxi into regions with higher long-term values. To get a better understanding of the cumulative reward, we randomly selected a location $l$ in uptown Manhattan and ran a full episode to generate the sequence of actions and rewards both for fixed policy and for DQN learned policy. We observed that during morning hours the DQN learned policy and fixed policy followed the same set of action sequences but later in the day, DQN learned policy started to compromise immediate rewards, and in turn, to get more long-term cumulative reward by forcing the taxi to move towards the high action-value regions.

\section{Conclusion}
\label{sec:conclusion}

We have developed a reinforcement learning system to generate an optimal carpooling policy for a taxi driver to maximize transportation efficiency in terms of fulfilling passenger orders. We have developed a carpool simulation environment using the historical taxi trip data to generate the samples of experience for training RL agent. To support an accurate simulator, we propose ST-NN, an end-to-end deep neural network approach that takes the raw GPS coordinates of origin and destination to estimate the travel time of potential trips. We conducted experiments on two different areas of Manhattan. The results show that the RL learned policy is able to intelligently decide when to accept a carpool trip based on the current driver state and the future prospect of the actions, with demonstrated advantage in optimizing the total effective trip distance of a driver within a day. In this work, we assume that decision-making for taxis are independent from each other.  One obvious future direction of research is to extend our framework to a multi-agent setting.  One other potential extension of this work is to make action space more granular to trip assignment. 

\bibliographystyle{IEEEtran}
\bibliography{ref_full}

\end{document}